\documentclass{article}

\usepackage{graphicx} %
\usepackage{titlesec}
\usepackage[margin=1.0in]{geometry}
\usepackage{float}
\usepackage{booktabs}
\usepackage{multirow}
\usepackage{graphicx}
\usepackage{amsmath}
\usepackage[T1]{fontenc}
\usepackage{lmodern}
\usepackage{microtype}
\usepackage{dsfont}
\usepackage{amssymb}
\usepackage{hyperref}
\usepackage{textcomp}
\usepackage{sidecap}
\hypersetup{
    colorlinks=true,
    linkcolor=blue,
    urlcolor=blue,
    citecolor=blue
}
\usepackage{siunitx}           
\usepackage[font=small,labelfont=bf]{caption} 
\usepackage{xparse}
\usepackage{xspace}
\usepackage{array}
\usepackage{ragged2e}
\usepackage{authblk} 
\usepackage[numbers,sort&compress]{natbib}

\usepackage[nolist]{acronym}

\makeatletter
\newcommand*{\org@overidelabel}{}
\let\org@overridelabel\@verridelabel
\@ifpackagelater{acronym}{2015/03/21}{%
  \renewcommand*{\@verridelabel}[1]{%
    \@bsphack
    \protected@write\@auxout{}{\string\AC@undonewlabel{#1@cref}}%
    \org@overridelabel{#1}%
    \@esphack
  }%
}{%
  \renewcommand*{\@verridelabel}[1]{%
    \@bsphack
    \protected@write\@auxout{}{\string\undonewlabel{#1@cref}}%
    \org@overridelabel{#1}%
    \@esphack
  }%
}
\makeatother

\titlespacing*{\section}
  {0pt}  %
  {2em}  %
  {2em}  %

\newcommand{\modelname}{Curiosity\xspace}

\newcommand{\eg}{e.g.\xspace}
\newcommand{\ie}{i.e.\xspace}

\title{Generative Medical Event Models Improve with Scale}

\author[1]{Shane Waxler\thanks{Co-first authors}}
\author[1]{Paul Blazek\protect\footnotemark[1]}
\author[1]{Davis White\protect\footnotemark[1]}
\author[1]{Daniel Sneider\protect\footnotemark[1]}

\author[1]{\\Kevin Chung}
\author[1]{Mani Nagarathnam}
\author[1]{Patrick Williams}
\author[1]{Hank Voeller}
\author[1]{Karen Wong}
\author[1]{Matthew Swanhorst}

\author[2]{Sheng Zhang}
\author[2]{Naoto Usuyama}
\author[2]{Cliff Wong}
\author[2]{Tristan Naumann}
\author[2]{Hoifung Poon}

\author[3]{Andrew Loza}
\author[3, 4]{Daniella Meeker}

\author[1]{Seth Hain}
\author[1]{Rahul Shah\thanks{\textit{Corresponding author: \href{mailto:rahul.shah@epic.com}{rahul@epic.com}}}}

\date{}

\affil[1]{Epic Systems}
\affil[2]{Microsoft Research}
\affil[3]{Yale School of Medicine}
\affil[4]{Cosmos Governing Council}

\begin{document}

\maketitle

\begin{acronym}
    \acro{ASCVD}{atherosclerotic cardiovascular disease}
    \acro{ATC}{Anatomical Therapeutic Chemical}
    \acro{AUCROC}{area under the curve of the receiver operating characteristic curve}
    \acro{CHF}{congestive heart failure}
    \acro{CKD}{chronic kidney disease}
    \acro{Comet}{Cosmos Medical Event Transformer}
    \acro{COPD}{chronic obstructive pulmonary disease}
    \acro{ECE}{expected calibration error}
    \acro{EHR}{electronic health record}
    \acro{HgbA1c}{hemoglobin A1c}
    \acro{HPB}{hepatopancreatobiliary}
    \acro{ICD-10-CM}{International Classification of Diseases, 10th Revision, Clinical Modification}
    \acro{LLM}{large language model}
    \acro{LOS}{length of stay}
    \acro{MAE}{mean absolute error}
    \acro{MCTD}{mixed connective tissue disease}
    \acro{NASH}{non-alcoholic steatohepatitis}
    \acro{NLP}{natural language processing}
    \acro{PM/DM}{polymyositis/dermatomyositis}
    \acro{PR-AUC}{precision-recall area under the curve}
    \acro{RMSE}{root-mean square error}
    \acro{RMSLE}{root-mean squared log error}
    \acro{RWD}{real-world data}
    \acro{RWE}{real-world evidence}
    \acro{SLE}{systemic lupus erythematosus}
    \acro{T2DM}{type 2 diabetes mellitus}
    \acro{TFLOP}{tera floating point operation}
    \acro{XGBoost}{gradient-boosted decision trees}
\end{acronym}

\begin{abstract}
Realizing personalized medicine at scale calls for methods that distill insights from longitudinal patient journeys, which can be viewed as a sequence of medical events. Foundation models pretrained on large-scale medical event data represent a promising direction for scaling real-world evidence generation and generalizing to diverse downstream tasks. Using Epic Cosmos, a dataset with medical events from de-identified longitudinal health records for 16.3 billion encounters over 300 million unique patient records from 310 health systems, we introduce the \modelname models, a family of decoder-only transformer models pretrained on 118 million patients representing 115 billion discrete medical events (151 billion tokens). We present the largest scaling-law study of medical event data, establishing a methodology for pretraining and revealing power-law scaling relationships for compute, tokens, and model size. Consequently, we pretrained a series of compute-optimal models with up to 1 billion parameters. Conditioned on a patient's real-world history, \modelname autoregressively predicts the next medical event to simulate patient health timelines. We studied 78 real-world tasks, including diagnosis prediction, disease prognosis, and healthcare operations. Remarkably for a foundation model with generic pretraining and simulation-based inference, \modelname generally outperformed or matched task-specific supervised models on these tasks, without requiring task-specific fine-tuning or few-shot examples. \modelname's predictive power consistently improves as the model and pretraining scale. Our results show that \modelname, a generative medical event foundation model, can effectively capture complex clinical dynamics, providing an extensible and generalizable framework to support clinical decision-making, streamline healthcare operations, and improve patient outcomes.
\end{abstract}

\acresetall %

\section{Introduction}
Safe and effective medical care aims to deliver the right intervention to the right patient at the right time. In pursuit of this goal, patients, clinicians, and health system leaders seek consensus-driven guidelines, integrated data sources, and richer information that captures the full diversity of real-world healthcare. Optimal health outcomes require care that excels across at least four pillars: accurate diagnosis, reliable prognosis, individualized treatment planning, and optimized clinical workflow~\cite{ml_in_medicine}. Succeeding across each of these pillars requires understanding a patient’s longitudinal medical history, addressing diagnostic and future uncertainty, incorporating patient values and goals, and adapting reasoning to temporal and clinical contexts.

\Ac{RWD} and \ac{RWE} offer a scalable path to personalized medical care. \ac{RWD}-driven insights already inform post-market safety surveillance, support regulatory approvals, and guide therapeutic strategies for complex chronic diseases~
\cite{sherman2016rwe, FDA_EHR_2018, concato2022rwe, usc2016cures}. Today, using \ac{RWD} to generate \ac{RWE} at scale demands significant analytic expertise and manual curation, constraining its day-to-day impact at the point of care~\cite{sherman2016rwe, concato2022rwe}. Unlocking its full potential will require methods that can transform raw data into actionable insights at the point of care in a scalable, generalizable, and personalized way.

Epic Cosmos\footnote{\url{https://cosmos.epic.com}} was created to address these challenges. A collaboration among health systems using Epic that is governed by a peer-elected council of participants, Cosmos aggregates de‑identified longitudinal health records for more than 300 million patients and 16.3 billion encounters as of August 2025, deduplicating each patient's records across health systems and combining them into a single integrated longitudinal record. This platform unifies common clinical data---including laboratory results, diagnoses, medications, and procedures---and includes other data relevant to health, such as social drivers of health, cancer staging, genomic variants, and patient-reported outcomes, among many other data types. The de-identified data in Cosmos is intended to support patient care and accelerate scientific discovery. Insights from Cosmos are delivered to clinicians today at the point of care through features in Epic like the Cosmos Median Length of Stay, Look-Alikes, and Best Care Choices for My Patient\texttrademark. Cosmos data and its downstream applications are only made available to health systems that contribute data to it. Cosmos has also been used to address a wide variety of research priorities~\cite{cosmosnih} such as understanding large trends in healthcare~\cite{2025Waves, 2024DecliningLamsal}, investigating rare diseases~\cite{Moskatel2025Prevalence, Kranyak2023Alopecia}, and analyzing healthcare utilization~\cite{opioidNoor2025, Turer2025Portals}.

Yet even at the scale of Cosmos, answering a single clinical question requires crafting custom cohort definitions, feature engineering pipelines, and statistical analyses. To enable personalized medicine and \ac{RWE} at scale for routine clinical decision-making, we need tools that can learn from the integrated patient record and flexibly answer complex medical inquiries, retrieving the right \ac{RWE} to support decisions across a wide variety of contexts.

Foundation models pretrained on real-world patient journeys have shown promise in addressing this problem, where a patient journey is formulated as a sequence of medical events. By learning latent representations of complete patient records, generative medical event models can provide patient‑specific predictions through simulated health timelines. By simulating multiple probabilistic timelines of a patient's health, quantitative predictions can be made about the likelihood of events over specific time intervals. A single set of generated trajectories can flexibly address a wide range of clinical queries even without prespecifying tasks, building task-specific models, or prompting a natural-language model with tailored questions. Furthermore, because medical foundation models are trained specifically on medical event tokens, they can be more parameter- and token-efficient in their representations. Generative medical event foundation models also offer extensive flexibility for forecasting future events, beyond binary or quantitative prediction tasks; for example, they can predict the most likely order of events or generate a set of all events within a time frame from among hundreds of thousands of possibilities. Medical foundation models also provide a unique form of interpretability, in that clinicians and researchers can study individual generated trajectories of medical events to better validate and understand the sequence of events that the model predicts may lead to downstream outcomes.

Previous models, such as CLMBR~\cite{stanford_clmbr}, MOTOR~\cite{stanford_motor}, Foresight~\cite{foresight}, ETHOS~\cite{ethos}, and others (see \autoref{sec:related_foundation}) have demonstrated the feasibility of this approach; however, their scopes are constrained by dataset breadth and depth, leaving the scalability of these approaches largely untested. 
Moreover, the choices of model size and compute have not been systematically studied, and it is unclear whether they are optimal and how they should scale with available data. This is particularly challenging for \ac{RWE} studies at the population scale of Cosmos, as suboptimal model size and compute can be extremely costly and wasteful.

To the best of our knowledge, \citet{msr_scaling} conducted the first comprehensive study of the scaling laws on structured patient records. They observed power-law relationships among compute, model size, and pretraining data similar to those in the text domain, albeit with a much higher optimal token-to-parameter ratio, which may be attributable to the distinct characteristics of medical events. In this paper, we apply the same methodology to pretraining on Cosmos data, producing to date the largest scaling-law study on real-world patient journeys. The study in \citet{msr_scaling} was limited to a de-identified dataset with only a few hundred thousand patients in emergency medicine. By contrast, our study is not only more than three orders of magnitude larger in patient count, but also covers an extremely diverse range of patient populations and health conditions.

We present \modelname, building on advances in generative medical event models by pretraining three decoder-only transformers with up to 1 billion parameters on Cosmos data. These models generate the next medical event---such as a diagnosis, medication order, lab result, the passage of time, or others---and these zero-shot generated sequences of medical events can be used to make clinical predictions. This work makes three contributions: 
\begin{enumerate}
    \item \textbf{\modelname models:} We describe the data transformation and training pipeline for medical event data on a dataset of 151 billion tokens derived from 115 billion medical events across 8.5 billion encounters.
    \item \textbf{Clinically relevant evaluations:} We show that zero-shot generations with \modelname models demonstrate strong predictive performance on a wide variety of clinical tasks.
    \item \textbf{Scalability:} We show that scaling up model and dataset size predictably decreases training loss and that minimizing train loss consistently improves downstream evaluation scores.
\end{enumerate}

By learning from the collective experience of care provided by the Cosmos community, \modelname captures patterns from data that are broad, rich, representative, and real. \modelname offers the potential for clinicians, researchers, and health systems to transform that experience into intelligence that uncovers new medical knowledge, enhances healthcare systems, and improves patient outcomes.

\section{Results}

\subsection{\modelname Training and Inference}

\begin{figure}[!htbp]
    \centering
    \includegraphics[width=\linewidth]{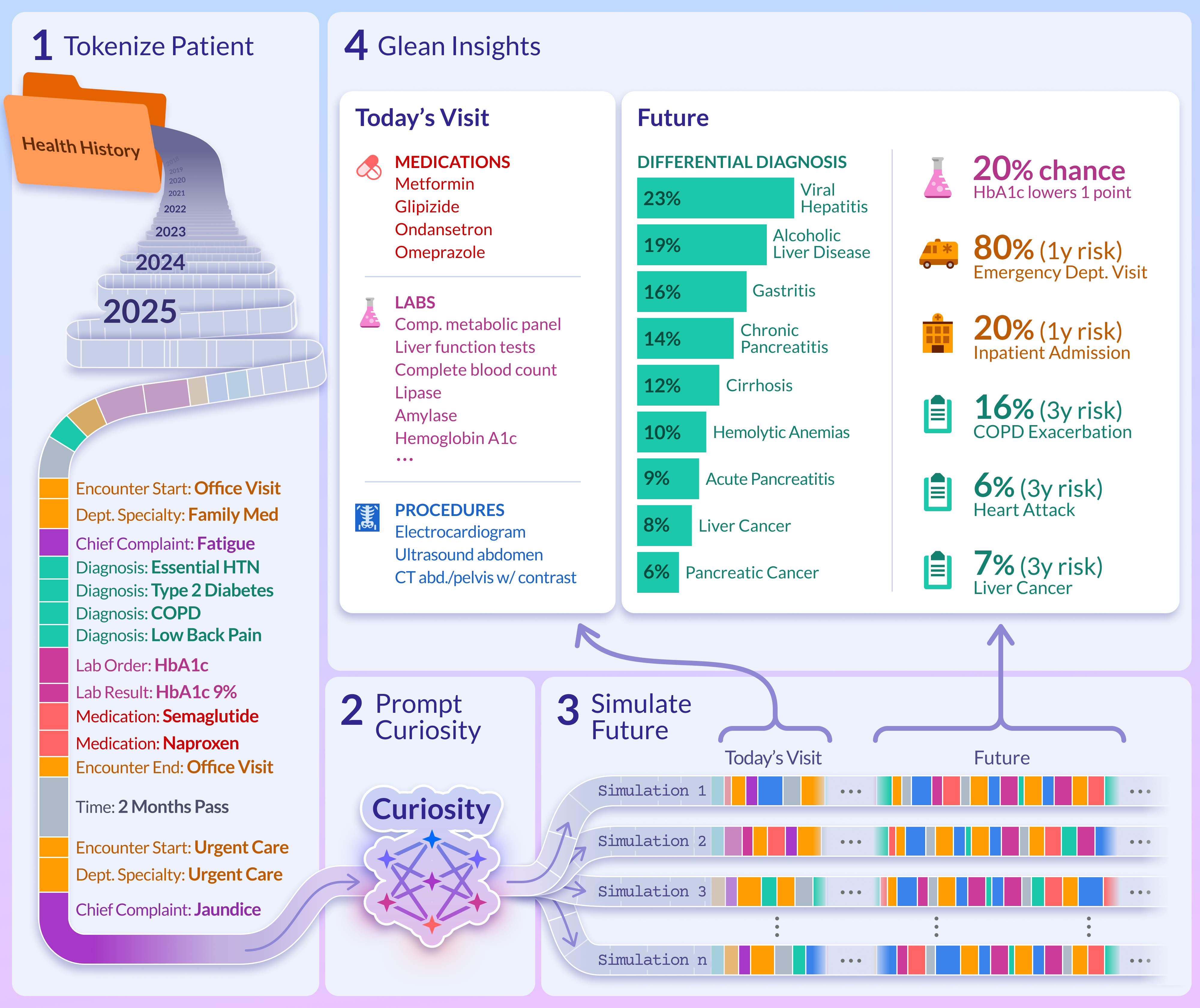}
    \caption[Model Inference]{\textbf{Overview of \modelname pretraining and inference.} 
    A patient journey is formulated as a sequence of medical events, and \modelname learns by predicting the next medical event. At inference time, \modelname is prompted with a patient's medical event history and simulates potential future trajectories by autoregressively generating the next events. Predictions for any target in \modelname's vocabulary are obtained from these simulated trajectories, enabling broad, out-of-the-box use on downstream tasks without task-specific fine-tuning or few-shot prompts.
    }
    \label{fig:prediction_pipeline}
\end{figure}

\label{sec:approach}
The dataset used for training and evaluating \modelname is a filtered subset of Cosmos that comprises 115 billion medical events from 118 million unique patient records spanning January 2012 to April 2025 (see \autoref{methods:data_preprocessing} and \autoref{tab:supp_datset_statistics}). We transformed each patient's medical events into a chronological sequence, where events are each represented by compact tokens. Certain tokenization methods were inspired by ETHOS~\cite{ethos} and adapted for the scale and heterogeneity of Cosmos data. \autoref{sec:methods} provides additional information on preprocessing, sequencing, and tokenization of Cosmos data. We trained \modelname using the Qwen2 transformer architecture~\cite{qwen2} with random initialization---\ie, without loading any pretrained Qwen2 weights (see \autoref{methods:training} for training details). Three model sizes were trained, as detailed in \autoref{tab:curiosity_models}. The optimal compute and training tokens used for each were determined by a scaling-law analysis, detailed in \autoref{sec:scaling_laws}.

\begin{table}[ht]
\centering
\small
\renewcommand{\arraystretch}{1.1}
\begin{tabular}{lcccc}
\toprule
\textbf{Name} & \textbf{Parameters} & \textbf{Training tokens} & \textbf{Compute} (\acsp{TFLOP})\\
\midrule
\modelname-S & 62M & 90B & 67M \\
\modelname-M & 119M & 160B & 234M \\
\modelname-L & 1B & 1.7T & 14B \\
\bottomrule
\end{tabular}
\caption{Trained compute-optimal \modelname models, with their size in parameters, number of training tokens, and amount of training compute measured in teraFLOPs (floating-point operations).}
\label{tab:curiosity_models}
\end{table}

\begin{figure}[!htbp]
    \centering
    \includegraphics[width=\linewidth]{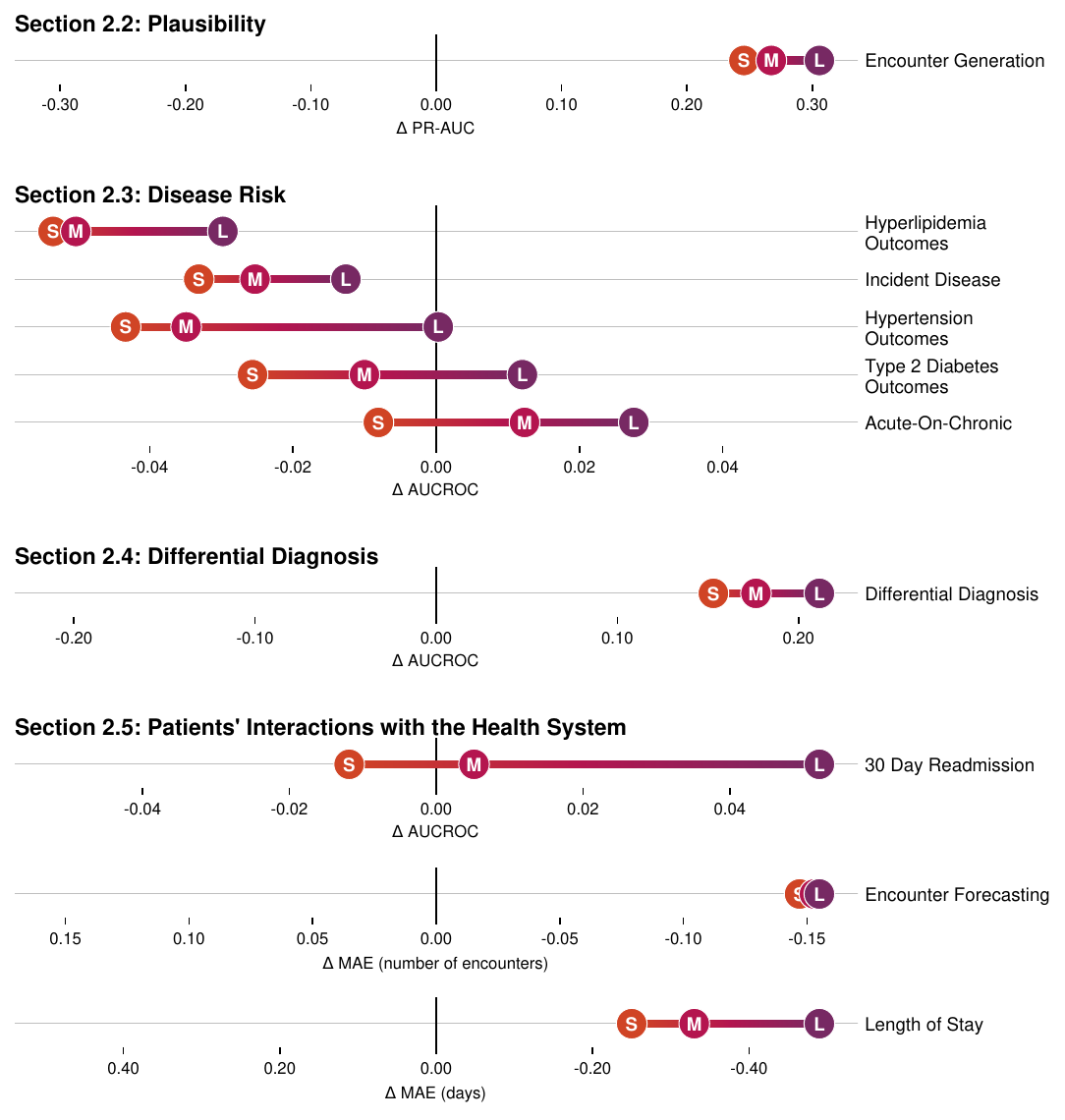}
    \caption[Summary statistics]{\textbf{Overview of \modelname evaluation performance.}  Each point shows the change in median evaluation scores for \modelname-S, \modelname-M, and \modelname-L relative to the best-performing task-specific supervised model in each of the major evaluation categories. For AUCROC and PR-AUC, positive values indicate that \modelname outperforms the task-specific model and negative values indicate underperformance while the opposite is true for MAE. \modelname's performance improved with scale and generally matched or even outperformed the best task-specific supervised methods.
    }
    \label{fig:summary_stat}
\end{figure}

\autoref{fig:prediction_pipeline} shows an overview of how \modelname inference works, and \autoref{fig:summary_stat} summarizes \modelname's performance across a wide range of clinically relevant evaluations. The model is prompted with a patient’s longitudinal, tokenized record up to a desired time point. \modelname probabilistically generates $n$ simulations, which are then used to compute all predictions, including event probabilities, distributions, times-to-event, and outcome collections (see \autoref{methods:inference} for more details). For all evaluation tasks listed below, models were evaluated on data from the test set (see dataset construction methods in \autoref{methods:data_preprocessing}). Full tables of evaluation results are in \autoref{sec:full_results}. \autoref{fig:summary_stat} provides a high-level snapshot of \modelname’s performance across all evaluation categories, with detailed task-level results and descriptions presented in the subsections that follow.

To contextualize \modelname's performance on these downstream tasks, we also trained three classes of supervised task-specific models (see \autoref{methods:supervised_baselines} for details): linear and logistic regression, \ac{XGBoost}, and supervised transformers trained from random initialization. Each of these task-specific models was trained on its corresponding downstream task and evaluated using the same datasets and procedures as those used for \modelname. For simplicity, figures only show the best-performing task-specific model.

\subsection{\modelname models generate realistic medical event sequences}
\label{sec:realism}
We evaluated alignment between a patient's ground truth health records and \modelname's generations to validate the plausibility of using these generations for more downstream predictive tasks.

\subsubsection{Plausibility statistics}
\label{sec:aggregate_plausibility}
We first examined aggregate summary statistics over \modelname generations, prompting \modelname to produce 25 1-year generations for 20,000 patients. We measured \modelname's ability to generate valid individual medical events that span multiple tokens, including diagnosis codes, medication codes, lab-result events, and department specialties in all encounter headers (see \autoref{sec:validity} and \autoref{tab:model_comparison_plausibility}). Generated multi-token events were rarely invalid, and the error rate decreased as model scale increased. Furthermore, the prevalence of individual medical events for diagnoses, medications, labs, and procedures within one year as generated by \modelname all strongly agreed with their corresponding prevalence in the same patients' 1-year future, as did the 1-year co-occurrence rate of pairs of medical events (see \autoref{sec:prevalence} and \autoref{tab:model_comparison_plausibility}).

\subsubsection{Encounter types and frequency}
\label{sec:utilization_calibration}

\begin{figure}[H]
    \centering
    \includegraphics[width=\linewidth]{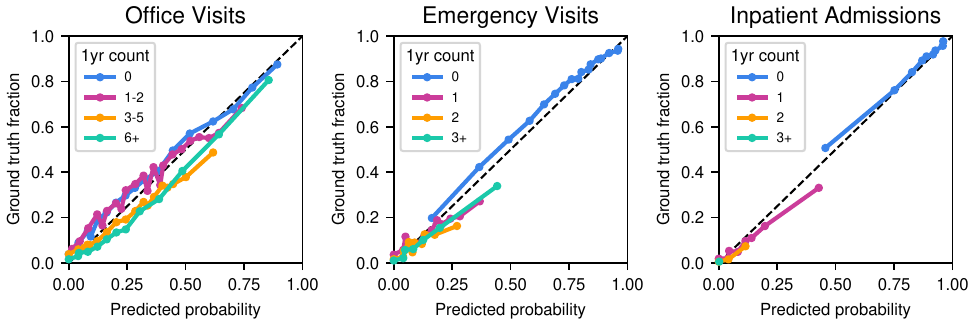}
    \caption[Utilization]{\textbf{Calibration plots for encounter frequency.} \modelname-L predicted the probability of how many encounters each patient will have within the next year, for three encounter types (Office Visit, Emergency, and Inpatient). Each point represents a quantile group containing an equal number of patients with similar predicted probabilities. The horizontal position of each point reflects the group's average predicted probability and the vertical position reflects the fraction of patients in that group with the specified 1-year count of encounters. Some lines do not span the full horizontal axis because few patients had those predicted probabilities. The diagonal line indicates perfect probability calibration.}
    \label{fig:utilizations}
\end{figure}

In addition to individual medical events, we measured how well \modelname probabilistically generates the number and types of medical encounters a given patient will experience within one year. Using the same model generations as \autoref{sec:aggregate_plausibility}, we found each patient's probability distribution for the number of office visits, emergency visits, and hospital admissions that would occur in the next year. In \autoref{fig:utilizations}, calibration curves for \modelname-L show good calibration for predicting personalized healthcare needs. Across encounter types and counts, the \ac{ECE}~\cite{Naeini_Cooper_Hauskrecht_2015} improved with model scale (\autoref{sec:full_results}).

\subsubsection{Single-encounter generations} %
\label{sec:encounter_generations}
We next measured how well \modelname generates the full set of unique diagnosis, medication, lab, and procedure events that will occur during an encounter. For three different encounter types (office visits, emergency visits, and inpatient admissions), we chose 10,000 random encounters and prompted \modelname with the patient's history up to and including the target encounter's header (\ie encounter type, department specialty, and any chief complaints). We compared the micro-averaged recall and precision of \modelname's encounter predictions to reference values representing the recall and precision of simply filling the encounter with the patient's past medical events over various lookback windows. 

\begin{figure}[!ht]
    \centering
    \includegraphics[width=\linewidth]{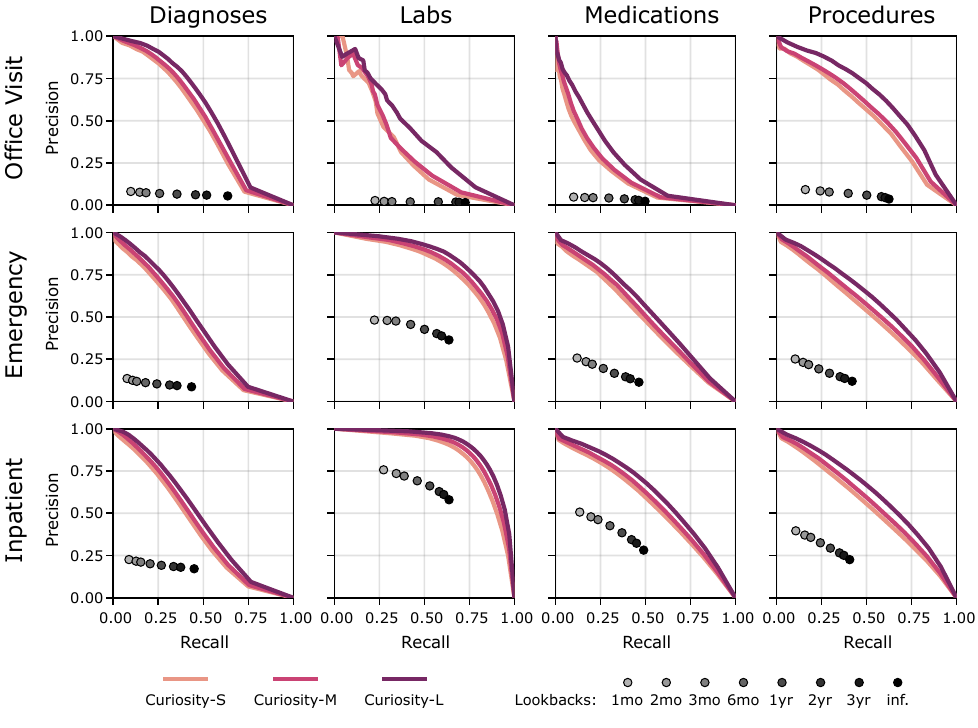}
    \caption[Encounter Generation]{\textbf{Medical events predicted for single encounters.} For office visit, emergency visit, and inpatient admissions, 10,000 random encounters of each were selected, and their medical events were compared to the medical events that \modelname predicted over 20 generations. The micro-averaged precisions and recalls are plotted over various thresholds for diagnosis, lab, medication, and procedure medical event types. In order to provide context on \modelname's performance, we pooled the patient's past events over various lookback windows and plotted the precision and recall for each. Higher area under each curve indicates better performance.}
    \label{fig:encounter_generation}
\end{figure}

\autoref{fig:encounter_generation} shows that across encounter types and medical event types, \modelname models demonstrated higher recall and precision than these lookback methods, and that with larger model sizes this performance showed consistent improvements as measured by \ac{PR-AUC} (see \autoref{tab:model_comparison_enc_gen}). These precision-recall curves indicate \modelname generated medical events the patient had not previously had prior to the current encounter \textit{and} filtered out most past events that were not likely to be repeated in the encounter.

\subsection{\modelname models can predict personalized future disease risk}
\label{sec:prognosis}

We investigated \modelname's ability to estimate future disease risk across multiple clinical domains and cohorts. Specifically, we categorized our prediction tasks into the following groups: disease-specific outcomes, acute-on-chronic events, and incident disease risk.

\subsubsection{Disease-Specific Outcome Predictions}
\label{sec:disease_outcomes}

We measured \modelname's performance on relevant disease-specific outcome prediction tasks, indexed to the time of care decisions. In particular, for \ac{T2DM}, hyperlipidemia (HLD), and hypertension (HTN), we examined \modelname's ability to predict patients' risk of relevant outcomes at the time of a change in medication management. Adverse outcomes, such as three-year stroke risk, were labeled as binary targets indicating whether the event occurred in the given time frame. Relevant lab results, such as \ac{HgbA1c} or total cholesterol, were labeled as binary targets at certain thresholds (\eg, \ac{HgbA1c} $<\!7$).

\begin{figure}[H]
    \centering
    \includegraphics[width=\linewidth]{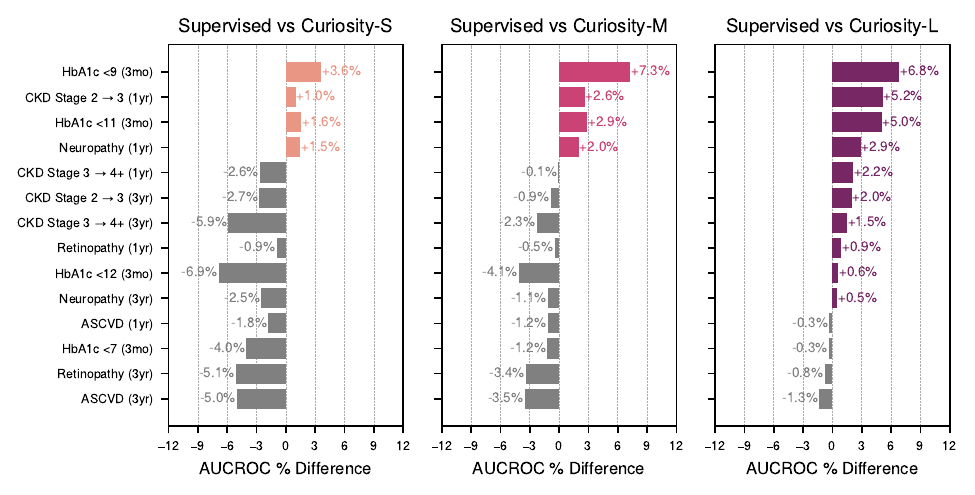}
    \caption[T2D]{\textbf{\ac{T2DM}-specific outcome predictions.} Percent increase of \ac{AUCROC} from the best-performing task-specific supervised model for each of the three \modelname models on the \ac{T2DM}-specific outcome prediction tasks.}
    \label{fig:T2D}
\end{figure}

\autoref{fig:T2D} illustrates \modelname's predictive performance on relevant outcomes for patients initiating a new medication therapy for active \ac{T2DM}. These include outcomes such as one-year and three-year risk for \ac{ASCVD}, \ac{CKD} progression from stage 2 through stage 4\textsuperscript{+}, diabetic neuropathy, and diabetic retinopathy, as well as two-to-four-month \ac{HgbA1c} lab results (the time frame when this lab is recommended to be reassessed after medication changes~\cite{a1c_cadence}). \modelname models improved consistently in these discriminative prediction tasks, with \modelname-L outperforming task-specific supervised models on most of these tasks. Scatter plots for predicted lab values can be found in \autoref{fig:dso_lab_plots}.

\begin{figure}[H]
    \centering
    \includegraphics[width=\linewidth]{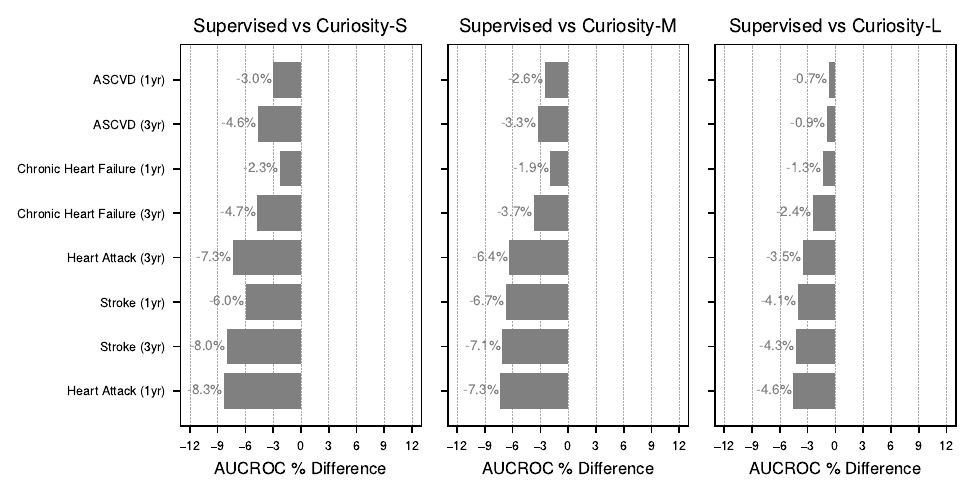}
    \caption[HLD]{\textbf{Hyperlipidemia-specific outcome predictions.} Percent increase of \ac{AUCROC} from the best-performing task-specific supervised model for each of the three \modelname models on the hyperlipidemia-specific outcome prediction tasks. \modelname models consistently scored better with scale, yet \modelname-L scored lower than the supervised models on each diagnosis task.}
    \label{fig:HLD}
\end{figure}

Likewise, \autoref{fig:HLD} shows results for \modelname predictions on tasks relevant to patients receiving treatment for hyperlipidemia, including one- and three-year risk of \ac{ASCVD}, heart attacks, strokes, and chronic heart failure (\eg, only the chronic diagnosis codes related to chronic heart failure). \modelname-L achieves an \ac{AUCROC} of 0.93 for predicting chronic heart failure diagnosis events within a year of changing hyperlipidemia medical management. While \modelname-L performance did not exceed the task-specific supervised models' performance, \modelname models showed measurable improvements as they scaled up, and absolute \ac{AUCROC} values were robust. Notably, \modelname did not outperform the task-specific models on the hyperlipidemia outcomes; the factors underlying this underperformance remain unclear and are left to future work.

\begin{figure}[H]
    \centering
    \includegraphics[width=\linewidth]{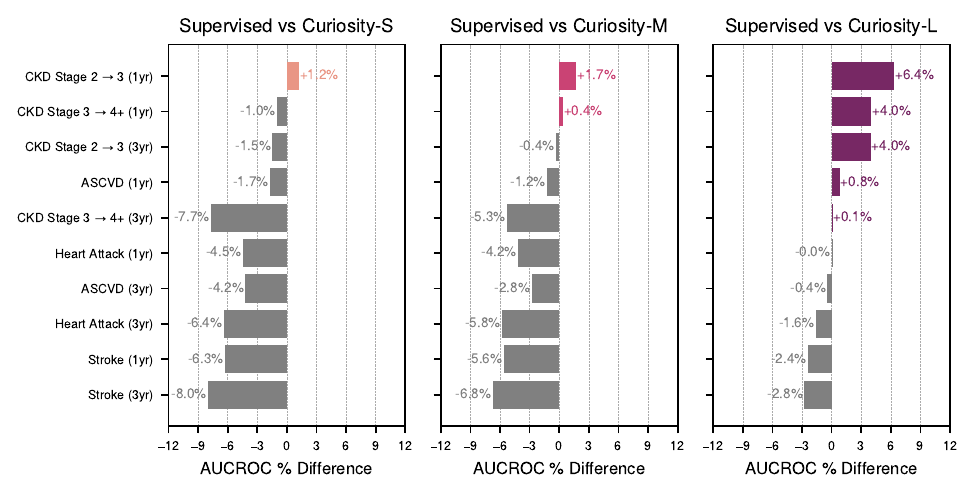}
    \caption[HTN]{\textbf{Hypertension-specific outcome predictions.} Percent increase of \ac{AUCROC} from the best-performing task-specific supervised model for each of the three \modelname models on the hypertension-specific outcome prediction tasks. \modelname-L scores matched or exceeded the supervised models on 6 out of 10 tasks.}
    \label{fig:HTN}
\end{figure}

For hypertension-related outcomes, we assessed \modelname's performances to make predictions about one- and three-year risk of \ac{ASCVD} events, heart attack, stroke, and \ac{CKD} progression from stage 2 to stage 3 and from stage 3 to stage 4\textsuperscript{+}. As with the two above cases, \modelname models improved as they increased in scale, with \modelname-L achieving higher \ac{AUCROC} scores than the task-specific models on half of these tasks. A full list of evaluation scores across disease-specific outcome tasks can be found in \autoref{tab:adverse_outcomes_restructured}.

\subsubsection{Acute-on-Chronic Incidence Prediction}
\label{sec:acute_on_chronic}

We next evaluated \modelname's ability to predict the two-year risk of acute-on-chronic clinical events, such as asthma exacerbation or sickle cell crisis. Patients are included in each acute-on-chronic cohort if they demonstrate a medical history of the relevant chronic disease (\eg, sickle cell disease for sickle cell crisis).

We formulated each acute-on-chronic evaluation as a binary classification task: for \ac{CHF} exacerbations for patients with chronic \ac{CHF}, asthma attacks for patients with asthma, sickle cell crises for patients with sickle cell disease, alcohol withdrawal syndrome for patients with alcohol use disorder, and \ac{COPD} exacerbations for patients with COPD. Detailed phenotype definitions, prediction date selection, and distinctions between chronic and acute event coding are provided in \autoref{methods:acute_on_chronic}. Approximately 5,000 unique patients are included for each group, with optional upsampling to ensure at least 500 patients experienced the acute event within two years of their prediction date. Dataset characteristics for each task can be found in \autoref{table:supp_eval_statistics}.

\begin{figure}[!ht]
    \centering
    \includegraphics[width=\linewidth]{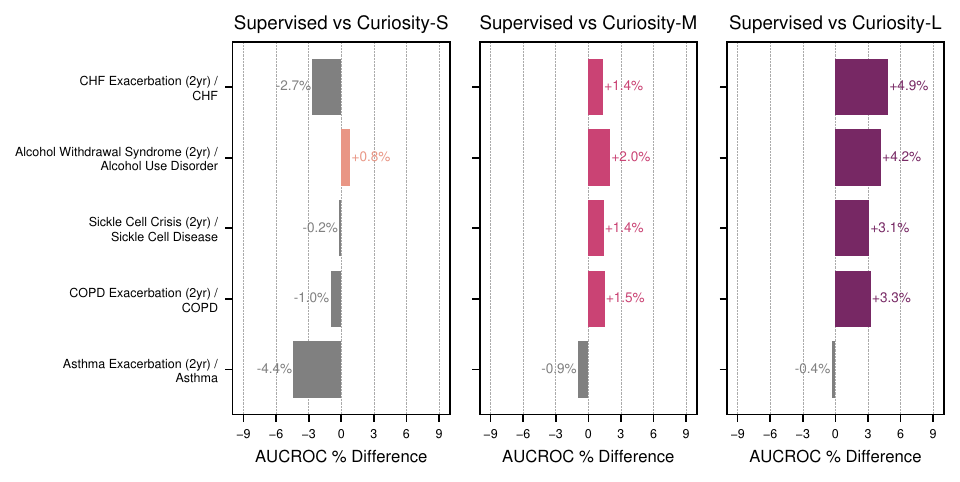}
    \caption[Acute-on-chronic]{\textbf{Acute-On-Chronic Tasks.} Percent increase of \ac{AUCROC} for each of the three \modelname models compared with the best-performing task-specific supervised model on the acute-on-chronic outcome prediction tasks. \modelname-M and \modelname-L scored higher than the baselines on 4 out of 5 tasks.}
    \label{fig:acute_on_chronic}
\end{figure}

\modelname-L achieved a higher \ac{AUCROC} than the task-specific models across all but one of these acute-on-chronic tasks, as shown in \autoref{fig:acute_on_chronic}, and in all tasks measured by \ac{PR-AUC}. All tasks showed consistently increasing \ac{AUCROC} and \ac{PR-AUC} with larger \modelname model size. Tabulated results can be found in \autoref{tab:roc-prauc-aoc}.

\subsubsection{Incident Disease Risk Prediction}
\label{sec:screening}

Finally, we tested \modelname's performance on predicting the first instance of a variety of disease states within a two-year period in the general population. \modelname generated 60 simulated timelines for approximately 5,000 patients, with each generation spanning two years of tokens. To avoid cohort biases and priors, broad inclusion and minimal exclusion criteria were used on these task cohorts (see \autoref{methods:population_screening} for details). Notably, this leads to class imbalance where all targets have a positive prevalence \textless1.5\%. This is quite different from the two preceding disease outcomes tasks, where the inclusion criteria for each task meant that those patients were naturally at increased risk for the prediction targets.

\begin{figure}[!ht]
    \centering
    \includegraphics[width=\linewidth]{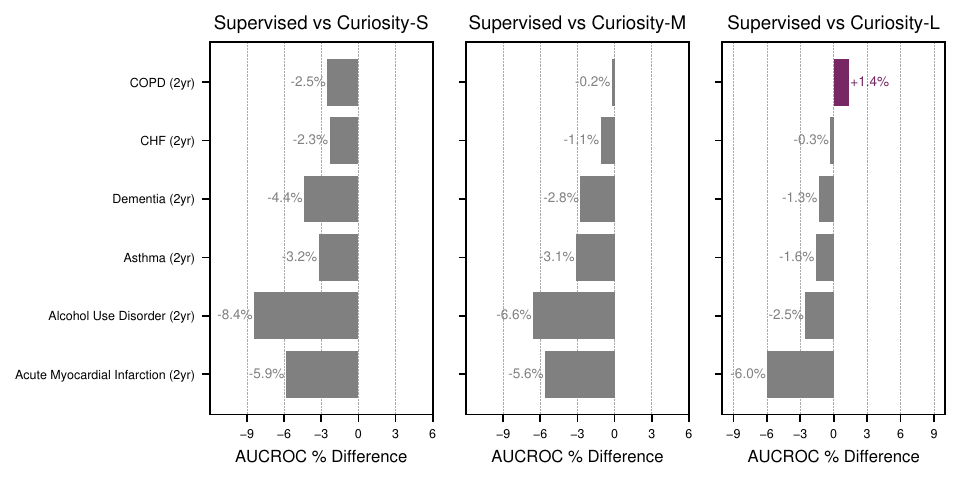}
    \caption[Screening]{\textbf{Incident Disease Risk Predictive Performance.} Percent increase of AUCROC from the best-performing task-specific supervised model for each of the three \modelname models on the six incident-disease prediction tasks. \modelname-L scores higher than the baselines on one out of six tasks.}
    \label{fig:screening}
\end{figure}

\autoref{fig:screening} compares \modelname to supervised models on incident disease tasks. \ac{AUCROC} scores are reported in the main figure for consistency with the other disease risk tasks. \modelname demonstrated higher \ac{AUCROC} scores than the supervised models on one out of six tasks and shows improvement across most tasks with model scale. However, in tasks like incident disease prediction with highly imbalanced class labels, \ac{PR-AUC} is more commonly used to judge performance~\cite{prauc_v_auroc,ozenne2015precision}. \modelname-L achieved higher \ac{PR-AUC} scores than the supervised models on all six tasks. Results for \modelname and supervised models (including \ac{PR-AUC}) are reported in \autoref{tab:roc-prauc-chronic}. One possible reason that \modelname-L did not outperform most task-specific models in AUCROC is because these tasks had extremely low prevalence and more generations per patient may be necessary. Preliminary evidence for this can be seen in \autoref{sec:ttc}.
  
\subsection[Diagnosis]{\modelname models generate early, quantitative differential diagnoses}
\label{sec:diagnosis}
We next assessed whether \modelname can provide ranked, quantitative differential diagnoses for individual patients. We selected two clusters of diagnoses, \ac{HPB} diseases and rheumatic diseases, because they span a variety of clinical presentations, can take multiple encounters to diagnose definitively, and have challenges with delayed diagnosis or misdiagnosis~\cite{rheumAAFP, rheumPatients, rheumEvaluation, hpbLiverEnzymes, hpbLiverCancerMisdiagnosis, hpbPancreaticCancerMisdiagnosis, hpbWhippleMisdiagnosis, hpbLiverDelayedDiagnosis}. Diseases within each cluster often present clinically with overlapping signs, symptoms, and laboratory findings, and as a result they often appear together on differential diagnoses. We selected nine diagnoses for both \ac{HPB} and rheumatic diseases, and for each we selected a cohort of patients who received that diagnosis (see \autoref{methods:differential_diagnosis} for diagnoses and their code-based definitions, and \autoref{tab:ddx_stats} for sample sizes). For each patient, we chose several points in their history in the one-year span before the first occurrence of their target diagnosis and had \modelname predict their risk for all the cluster's diagnoses. In this way, \modelname produced a ranked, quantitative differential diagnosis at multiple time points for each patient and flagged patients at risk of receiving these diagnoses. We did not prompt \modelname with any information beyond the start of the encounter at which the patient first received their target diagnosis, so \modelname never sees any diagnostic workup or documentation from this encounter.

\begin{figure}[!ht]
    \centering
    \includegraphics[width=1\linewidth]{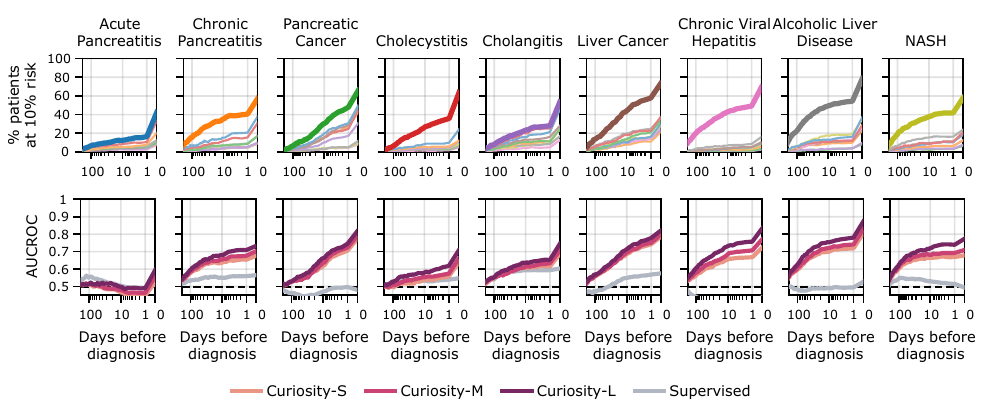}
    \caption[Early differential diagnosis]{\textbf{Hepatopancreatobiliary differential diagnosis.} \modelname-L was evaluated on predicting 1.5-year risk of receiving one of the indicated \ac{HPB} diagnoses. Each column represents a cohort of patients who were ultimately diagnosed with the indicated diagnosis. Each line in the first row represents the percentage of that cohort that was flagged by \modelname-L as having at least 10\% risk of a diagnosis. The correct diagnosis is shown in bold, the off-target diagnoses are faint; each line color represents the same diagnosis across the row. The second row shows the \ac{AUCROC} over time for all three \modelname models and the task-specific supervised model for predicting the indicated diagnosis.}
    \label{fig:hepato_diff_diagnosis}
\end{figure}

\begin{figure}[!ht]
    \centering
    \includegraphics[width=1\linewidth]{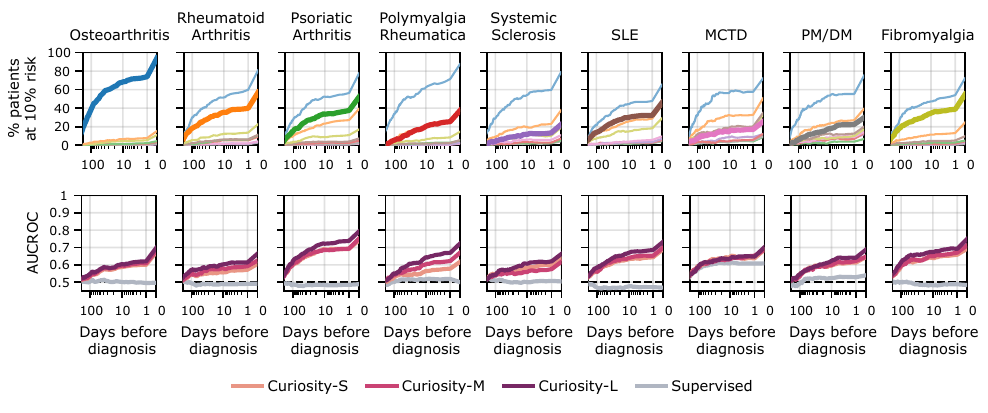}
    \caption[Early differential diagnosis]{\textbf{Rheumatic differential diagnosis.} \modelname-L was evaluated on predicting 1.5-year risk of receiving one of the indicated rheumatic diagnoses. Each column represents a cohort of patients who were ultimately diagnosed with the indicated diagnosis. Each line in the first row represents the percentage of that cohort that was flagged by \modelname-L as having at least 10\% risk of a diagnosis. The correct diagnosis is shown in bold, the off-target diagnoses are faint; each line color represents the same diagnosis across the row. The second row shows the \ac{AUCROC} over time for all three \modelname models and the task-specific supervised model for predicting the indicated diagnosis.}
    \label{fig:rheum_diff_diagnosis}
\end{figure}

Each plot in the top row of \autoref{fig:hepato_diff_diagnosis} and \autoref{fig:rheum_diff_diagnosis} shows how many patients were flagged by \modelname-L as having at least 10\% risk for the target diagnosis (bold) and each of the off-target diagnoses within the cluster (lighter, thinner lines). Each plot therefore represents the average differential diagnosis for the cohort over time. For the majority of diagnoses, \modelname-L correctly flagged more than 50\% of patients for their target diagnosis by their final prediction time. For most diagnoses, \modelname-L also flagged more than 25\% of patients weeks ahead of time.

For comparison, we trained task-specific supervised models to predict each of these diagnoses in a wider, more general pool of patients (see \autoref{methods:supervised_baselines} for details on training). These task-specific models were generally not able to differentiate well among patients within these narrower \ac{HPB} or rheumatic diagnosis clusters, with \ac{AUCROC} metrics not increasing appreciably over time (see the bottom row of both \autoref{fig:hepato_diff_diagnosis} and \autoref{fig:rheum_diff_diagnosis}), despite achieving modest sensitivity. In contrast, \ac{AUCROC} metrics from \modelname generally increase over time and as model size increases. This suggests that \modelname models can effectively distinguish patients with similar presentations but different eventual diagnoses, and they can produce realistic differential diagnoses that become more sensitive and specific as the patient's clinical presentation and diagnostic workup evolve. \ac{AUCROC} at the final prediction time is reported for all diagnoses and models in \autoref{tab:ddx_results}. Of note, these analyses do not address whether early diagnosis flagging reflects the diagnostic workup being pursued by the patient's medical providers \textit{versus} the ability to preemptively flag diagnoses before significant clinical suspicion.

\autoref{fig:hepato_diff_diagnosis} shows that, among all nine \ac{HPB} diseases, the correct diagnosis was the one most commonly flagged by \modelname-L, with the gap between the first- and second-ranked diagnoses generally increasing with time. For acute pancreatitis, \ac{AUCROC} is hardly better than chance until the target diagnosis date. For diseases with generally more insidious onset (\eg, cancer, chronic viral hepatitis, and alcoholic liver disease), \modelname-L was able to flag many patients much earlier (\autoref{fig:hepato_diff_diagnosis}).

For rheumatic disorders (\autoref{fig:rheum_diff_diagnosis}), most patients were flagged at the 10\% risk levels for osteoarthritis at some point, regardless of their eventual diagnosis. This is likely both because osteoarthritis is a common diagnosis for patients with undifferentiated joint pain and because of early inaccurate or imprecise diagnosis and documentation. Among the remaining eight target diagnoses, the correct diagnosis was ranked second for five cohorts, third for two, and fourth for one. \ac{AUCROC} scores for \modelname models ranged from 0.66-0.79 across rheumatic diagnoses at the final prediction time (\autoref{tab:ddx_results}).

\subsection{\modelname models forecast patients' interactions with the health system}
\label{sec:utilization}
Reliably forecasting health system interactions enables clinicians and health systems to plan for the needs of their patients. We assessed \modelname's generated patient timelines for their ability to make accurate predictions about a patient's interactions with the healthcare system.

\begin{figure}[H]
    \centering
    \includegraphics[width=0.9\linewidth]{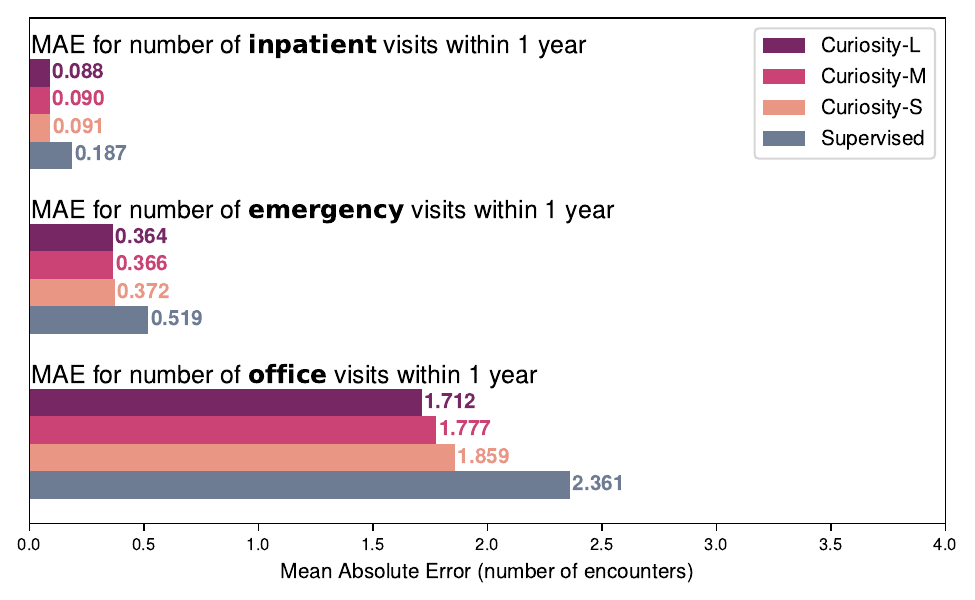}
    \caption[Encounter frequency comparison]{\textbf{One-year encounter frequency forecasting.} \modelname models were compared with the best-performing supervised task-specific regression model for predicting the number of inpatient, emergency, and outpatient encounters that will occur within a year's time for 18,400 patients. Mean absolute error (MAE) was used to measure the error in predicting these encounter counts; smaller is better.}
    \label{fig:combined_util}
\end{figure}

Having earlier assessed the probability calibration of predicting the number of encounters over a year in \autoref{fig:utilizations}, we next asked how close these predictions were to the ground truth number of encounters. For inpatient, outpatient, and emergency encounters, all three \modelname models demonstrated lower \ac{MAE} than supervised task-specific regression models for predicting future encounter counts. Results are shown in \autoref{fig:combined_util}.

\subsubsection{Hospital length of stay and 30-day readmission}
\label{sec:los_readmissions}
Accurate hospital \ac{LOS} prediction helps health systems manage beds and plan patient care to limit discharge delays~\cite{los_justification}. We evaluated \modelname's ability to predict \ac{LOS} on 10,000 randomly selected hospital admissions (see \autoref{tab:los_stats} for evaluation set statistics). All models received the patient's history through the admission encounter's header (\ie, encounter type, department specialty, and any chief complaints), in addition to the documented primary diagnosis.

\begin{figure}[H]
  \centering
  \begin{minipage}[c]{0.55\textwidth}
    \includegraphics[width=\linewidth]{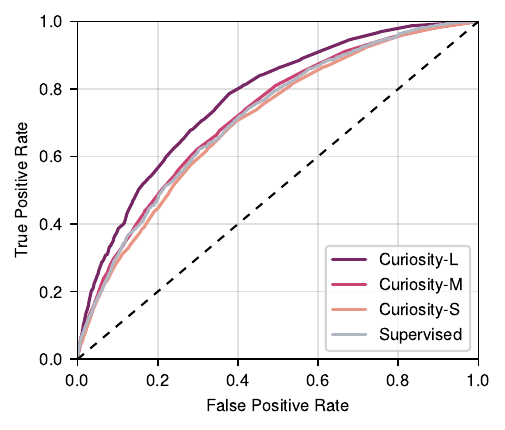}
  \end{minipage}%
  \hfill
  \begin{minipage}[c]{0.4\textwidth}
    \vspace*{\fill}
    \captionof{figure}{
      \textbf{30-day readmission prediction.}
      ROC curves for predicting hospital readmission within 30 days of discharge across 10,000 hospital encounters. \modelname-L achieved an \ac{AUCROC} of 0.770, with the best-performing supervised task-specific model achieving 0.717.
    }
    \label{fig:30_day_readmission}
    \vspace*{\fill}
  \end{minipage}
\end{figure}

Another metric for assessing healthcare utilization is the 30-day hospital readmission risk, an important and well-studied operational consideration for care transitions, discharge planning, and outpatient follow-up~\cite{readmission}. We randomly selected 10,000 patients being discharged from the hospital and had models predict the probability that a new hospitalization would occur within 30 days (patients who were readmitted within one day were excluded from this analysis because they may often reflect hospital transfers or clerically erroneous discharges \cite{cms2024HWR}). \modelname models saw increases in \ac{AUCROC} on this task, with \modelname-L and \modelname-M demonstrating higher scores than task-specific supervised models (see \autoref{fig:30_day_readmission} and \autoref{sec:full_results}).

\begin{figure}[H]
    \centering
    \includegraphics[width=0.9\linewidth]{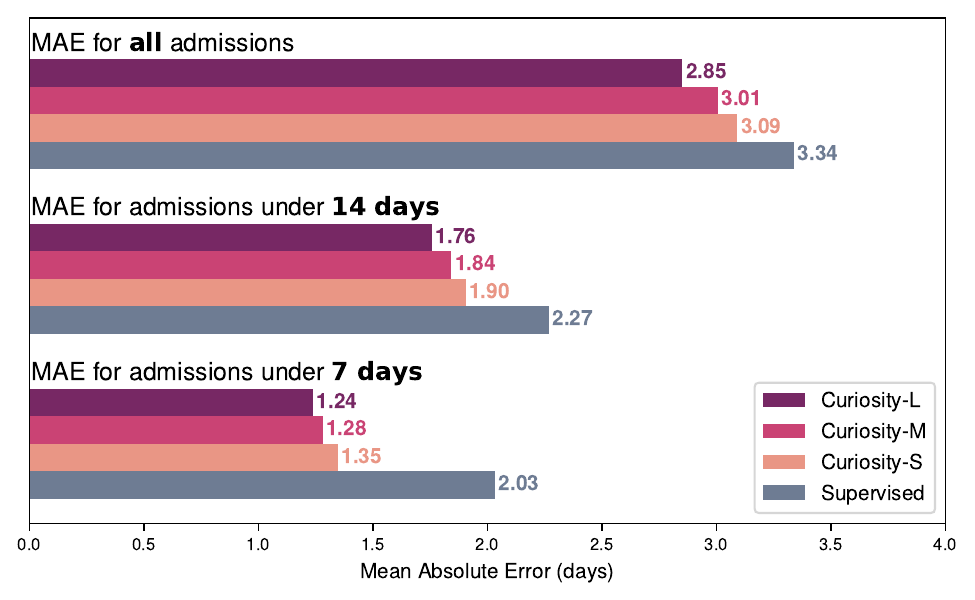}
    \label{fig:length_of_stay}
    \caption[Comparison of images]{\textbf{Hospital length of stay prediction.} Each of the \modelname models and the best-performing task-specific model evaluated on \ac{LOS} prediction on a set of 10,000 randomly selected samples. Mean absolute error (MAE) of the length of stay in days was used to compare model performance; smaller is better.}
    \label{fig:combined_los}
\end{figure}

\subsection{Training medical event models follows scaling laws}
\label{sec:scaling_laws}
Before training \modelname-S, \modelname-M, and \modelname-L, we first trained many smaller models to derive scaling laws \cite{chinchilla, scalinglawsopenai} to predict the optimal model size and number of training tokens for a given compute budget, measured in \acp{TFLOP}. This step was important not only for understanding the best parameters for training models, but also to understand how to optimally scale medical event foundation models on a sufficiently large dataset. Building on recent work demonstrating power-law scaling for generative medical event prediction on the MIMIC-IV dataset~\cite{msr_scaling}, we applied the same approach to Cosmos by training a sweep of 10 model sizes ranging from two million to one billion parameters on our dataset of over 136 billion training tokens.

\begin{figure}[!htbp]
    \centering
    \includegraphics[width=\linewidth]{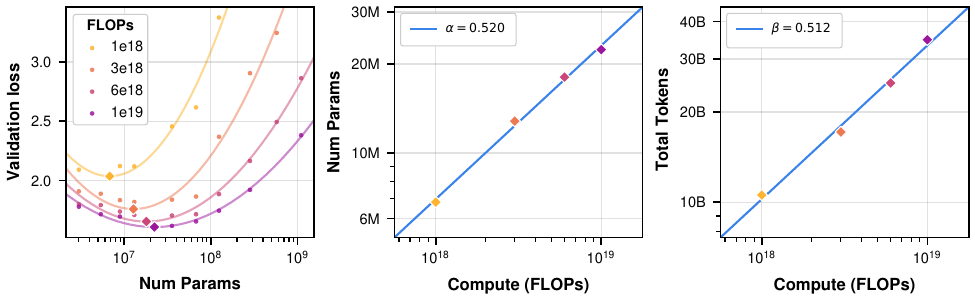}
    \caption[Scaling Laws]{\textbf{Optimal training of \modelname models follows power laws.} The minimum loss achieved by training runs with fixed compute and varied number of model parameters \textbf{(left)}. A log-scale parabola is fit to each isoFLOP curve, and the minimum point of each is marked by a diamond. \textbf{Middle} and \textbf{right} show how the isoFLOP parabolas' minima $N_{\text{opt}}$ and $D_{\text{opt}}$, respectively, vary with the isoFLOP compute on a log-log scale with power-law fits.}
    \label{fig:scaling_laws}
\end{figure}

As in \citet{chinchilla}, we ran a grid search over varying amounts of training \acp{TFLOP} to find the optimal scaling of model size and training tokens for the \modelname medical event dataset. After performing smaller training runs at four fixed compute budgets, we fit parabolas to the isoFLOP curves (\autoref{fig:scaling_laws}). 

The minima of these curves were used to fit power-law equations for the compute-optimal scaling of parameter count $N_\text{opt}$ and training tokens $D_\text{opt}$ with respect to the amount of compute $C$ used during training (\autoref{fig:scaling_laws}). Our experiments showed that for power laws of the form
\begin{equation}
\begin{aligned}
N_\text{opt} &= AC^{\alpha} \\
D_\text{opt} &= BC^{\beta}
\end{aligned}
\end{equation}

with fit parameters $A$, $B$, $\alpha$, and $\beta$, we obtained best-fit parameters of $\alpha = 0.520$ and $\beta = 0.512$, closely mirroring coefficients reported for natural‑language corpora ($\alpha = 0.49$ and $\beta = 0.51$)~\cite{chinchilla}. The near-equality of $\alpha$ and $\beta$ implies that, on the compute-optimal frontier, model size and training-token count should be scaled proportionally as total compute increases. We used these best-fit lines to derive the training parameters to train our 3 compute-optimal \modelname models, listed in \autoref{tab:curiosity_models}.

\subsection{Scaling medical event model performance}
\subsubsection{Performance vs. loss}
\label{sec:perf_v_loss}
Our scaling analysis demonstrated that increasing parameters and training tokens predictably decreases model loss. This raises the question of how minimized loss corresponds with downstream, clinically relevant evaluations. To measure this relationship, we evaluated the \modelname models and several training checkpoints at different train loss on the single-encounter generation, \ac{T2DM}-specific outcomes, and 30-day readmissions tasks.

\begin{figure}[H]
    \centering
    \includegraphics[width=\linewidth]{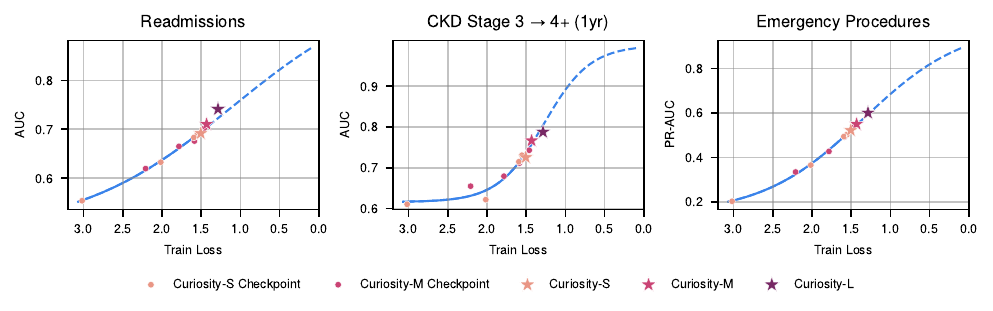}
    \caption[Performance vs Loss]{\textbf{Downstream performance improves as train loss decreases.} We evaluated each of the \modelname models, as well as earlier checkpoints from the \modelname-S and \modelname-M training runs, on a variety of our downstream evaluations. We fit a sigmoid curve to all points except those from \modelname-L to assess the sigmoid curve's predictive utility. We evaluated all of these models using a more conservative $n=20$ simulations.}
    \label{fig:performance_vs_loss}
\end{figure}

\autoref{fig:performance_vs_loss} shows a smooth, sigmoidal relationship between the training loss and downstream clinical evaluation scores (the full set of plots can be found in \autoref{fig:t2d_perf_v_loss}, and \autoref{fig:el_perf_v_loss}). This relationship holds over many different model sizes and compute budgets. The empirically fitted sigmoid functions show different inflection points and relative slopes, indicating that different clinical tasks require different levels of training to show improvements.

\subsubsection{Scaling test-time compute}
\label{sec:ttc}
At inference time, one of the uniquely relevant parameters for \modelname is the number of generated patient timelines, $n$. Because probabilities are calculated as an aggregation over $n$ generations (\autoref{methods:inference}), increasing $n$ scales inference costs linearly while reducing the Monte Carlo variance of downstream predictions and increasing the resolution of output probabilities. Unlike in language models, where test-time compute navigates a fully expressive space, in our setting, increasing $n$ also expands the method's expressiveness, raising the performance ceiling by reducing the quantization of predictions. To investigate this tradeoff, we varied $n$ for three representative clinical tasks: hyperlipidemia-specific outcomes (an evaluation with relatively high-prevalence positive outcomes), incident disease prediction (an evaluation with the lowest-prevalence positive outcomes), and single-encounter emergency generations (which is highly multi-target and uses \ac{PR-AUC}).

\begin{figure}[!ht]
    \centering
    \includegraphics[width=\linewidth]{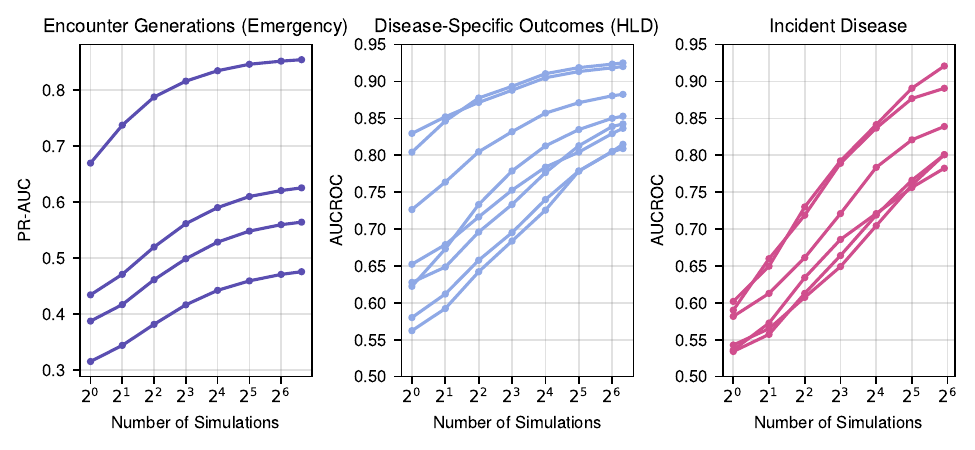}
    \caption[Scaling test-time compute]{\textbf{Effect of test-time compute on performance.} For \modelname-L we evaluated model performance against the number of simulations generated, focusing on single-encounter generations for emergency visits (\textbf{left}), HLD-specific outcomes (\textbf{middle}), and incident disease tasks (\textbf{right}). Each line represents a specific task (\eg, one-year \ac{ASCVD} risk, two-year COPD, etc.) from the titled category. For readability the legend is not included.}
    \label{fig:ttc}
\end{figure}
All three task groups in \autoref{fig:ttc} show a steady increase in performance as $n$ increases, with various degrees of plateauing at $n \gtrsim 64$. The incident disease evaluations are particularly sensitive to $n$, possibly due to most patients having a low probability of having these events.
Similar to language models~\cite{snell2024scalingllmtesttimecompute}, increasing train and test-time compute provide orthogonal directions to improve performance on downstream tasks.

\section{Discussion}
\label{sec:discussion}
\subsection{Key findings}
Here we demonstrate that large‐scale, medical event foundation models can learn from longitudinal patient records to produce realistic and useful clinical predictions. Trained on a subset of the 16.3B encounters across 300M patient records as of August 2025 from Cosmos, \modelname is a family of decoder-only transformer models up to 1B parameters in size that probabilistically generate medical event sequences which not only preserve event- and encounter-level realism but also demonstrate a broad range of short- and long-term predictive capabilities.

Previous medical event foundation models have been constrained by training on smaller datasets or a limited number of data types \cite{shakyllm}. \modelname models were trained directly on sequences of structured medical events spanning core clinical data types (\ie, demographics, encounters, diagnoses, chief complaints, labs, medications, procedures, and time). Rather than relying on natural language inputs and outputs, \modelname directly receives discrete medical events from a patient's medical record as input, learning event-level structure, long-range temporal dependencies, and the underlying probability distribution of medical events. As of this writing, \modelname represents the largest medical-event foundation model by number of medical events used for training.

Evaluation of medical event foundation models has typically been performed on relatively few tasks \cite{shakyllm}. By assessing \modelname's performance across a broad and diverse suite of clinical tasks spanning single-encounter generation, disease-specific outcome prediction, acute-on-chronic risk estimation, incident disease detection, differential diagnosis, and operational forecasting, we explored its potential as a flexible and generalizable tool for a range of clinical contexts. \modelname-L, our largest model, matched or outperformed supervised task-specific models on most of these tasks. Performance gains were evident across multiple clinical domains, supporting \modelname's generalizability and utility for diverse downstream applications.

We also conducted the largest scaling law analysis for medical event data to date---more than $300\times$ larger than the largest previous work \cite{msr_scaling}. To investigate medical event scaling laws, we employed the same methodology from \cite{chinchilla}, exploring isoFLOPs across four fixed compute points. Our isoFLOPs analysis yielded similar exponents to those shown in the natural language (NL) domain ($\alpha=0.520$ and $\beta=0.512$), suggesting we should scale training tokens and model parameters proportionally to train compute-optimal models \cite{chinchilla}. In contrast to work in the NL domain, we observed an optimal token-to-parameter ratio of ~1,000:1, closely matching results from prior work on medical event scaling laws \cite{msr_scaling}.

Lastly, we found that loss minimization predictably translates into better downstream performance on our suite of evaluations. Nearly every clinical task we evaluated exhibited sigmoidal improvements in its performance metrics as we minimized training loss. These trends were consistent across task families and clinical domains, and most tasks had not yet plateaued, indicating substantial remaining room for improvement. This result highlights pretraining loss as a useful proxy for downstream capability, and suggests that continuing to scale will yield better downstream utility. In addition to training-time scaling, we also observe benefits from increased inference-time compute: generating more patient simulations per prediction consistently improves downstream performance metrics across many clinical tasks.

\subsection{Limitations}
\label{discussion:limitations}
Even with these strengths, \modelname has clear limitations.

\textbf{Real-world data.} \modelname is trained on real-world healthcare data in Cosmos, and it is subject to imprecision and errors in documentation and clinical practice~\cite{liu2022real}. Several automated data quality control processes in Cosmos aim to improve data completeness and identify irregularities, and these periodic analyses are fed back to the contributing organizations~\cite{cosmosnih}. Cosmos aggregates data across 310 health systems with extensive linking and deduplication efforts, but individuals may receive care at other health systems that is not recorded in Cosmos. This leads to data gaps before, within, and after observed medical event sequences. Certain prediction targets may be missing or occur earlier than the first record in Cosmos. We mitigate this by training and evaluating on patients who meet predefined criteria for observable healthcare utilization (see \autoref{methods:data_preprocessing}) and by using time tokens, which allow the model to read and generate patterns of absent data. 

\textbf{Tokenization}. While the tokenizer for \modelname can be used to encode a wide range of clinical data, the discretization of continuous values like laboratory results and time masks smaller changes in these values that may be important for some prediction tasks. Future work can investigate trade-offs in performance and model complexity between a pure categorical approach and a hybrid approach that handles numeric values separately~\cite{2023eventstreamgptdata, 2024xvalcontinuousnumericaltokenization, 2025multivariategptdecoderonlytransformermultivariate, 2023embeddingsnumericalfeaturestabular, 2021tabulartransformersmodelingmultivariate}. In addition, we did not conduct an evaluation for training a tokenizer or optimizing vocabulary size, presenting a rich avenue for future work. Tokenizers trade vocabulary size for sequence length, which can impact performance in nontrivial ways with model scale~\cite{tao2024scalinglawsvocabularylarger}. As shown in \autoref{fig:TokenDist}, many patients' entire tokenized history exceeded the model's current context window of 8,192, presenting an opportunity to improve performance with longer context models. Finally, while \modelname tokenized and was trained on a core set of structured clinical data types \autoref{methods:tokenization}, many other data types remain to be added.

\textbf{Evaluation}. Disease phenotype definitions were based on \ac{ICD-10-CM} codes, categorical encounter types, demographics, event timings, and other similar features. While these features form reasonable phenotypes, they do not always correspond directly to how events transpired in the real world. This presents an opportunity for future work involving prospective validation of \modelname's outputs. Finally, evaluations focused on aggregate performance across individuals in the test set and did not investigate performance and calibration within specific subpopulations, which will be an additional area for future work.

\subsection{Future work}
The above constraints outline the frontier of \modelname's capabilities. However, each constraint naturally suggests consequent experiments and improvements.

\textbf{New event types.} First, including additional structured data from Cosmos, such as genomic variants, social drivers of health, and cancer staging data, would give \modelname a more robust representation of patient health timelines. The addition of specialty-focused data types would further improve \modelname's ability to both contextualize and make certain predictions for a wide range of medical specialties, professions, and use cases. Pediatric data, including mother-baby links, could enable \modelname to capture developmental physiology, age-specific drug dosing, vaccine schedules, and congenital condition timelines. Lastly, multimodal data such as waveform vitals, images, and clinical notes would close information gaps that structured medical event data cannot capture.

\textbf{Counterfactual reasoning.} Future work will incorporate decision-effect estimation objectives and reinforce counterfactual consistency; explicit what-if analyses (\eg, ``What if metformin were initiated today?'') will allow us to move beyond risk analysis and toward actionable risk mitigation.

\textbf{Time-to-event analysis.} Our evaluation tasks focused primarily on binary predictive outcomes, and censoring was not explicitly accounted for in our analysis of real and synthetic patient timelines. This does not take advantage of the full time-to-event predictive capacity that \modelname has to offer.

\textbf{Fine-tuning.} While \modelname achieves strong zero-shot performance, it can also be fine-tuned to improve task-specific performance and extend its capabilities to out of vocabulary tasks. Fine-tuning offers an alternative to increasing test-time compute, eliminating sampling variance and enabling multiple precise calibrated predictions to be made with a single compute-efficient forward pass.

\textbf{Prospective studies.} Prospective studies could help assess how well \modelname outputs align with clinical practice, which would inform downstream applications and model retraining schedules. 

\textbf{Human factors research and governance.} \modelname models are generalizable to many clinical tasks, and deeply understanding best practices for introducing and scaling downstream applications within clinical workflows is key to interpretability, transparency, and usability. As with other healthcare technology, strong evaluation frameworks and thoughtful governance strategies are important to make \modelname applications sustainable, responsible, and fair.
\paragraph{}
Together, these findings position \modelname as a general‐purpose engine for real-world evidence: it can screen populations for incident disease risk, forecast individual health timelines, surface differential diagnoses, and anticipate health system interactions---all from a single set of autoregressive generations.

\section{Related Work}
\label{sec:related_work}
In this section, we highlight prior work that informs and contextualizes our study, focusing on three key areas: (1) large-scale, real-world medical event databases such as Epic Cosmos, (2) scaling laws for large language models, and (3) medical foundation models trained on structured and unstructured health data. This is not an exhaustive survey of related literature; rather, we summarize representative work that captures the state of the art, key challenges, and motivations for our approach.

\subsection{Epic Cosmos}
As described above, Cosmos is a collaboration for integrating and linking \ac{EHR} data among 310 healthcare organizations\footnote{\url{https://cosmos.epic.com/community}} in the United States, Canada, Lebanon, and Saudi Arabia as of August 2025~\cite{cosmosnih}. The Cosmos population from healthcare organizations in the United States is largely representative of U.S. census data. While some variation from national statistics has been noted, Cosmos has been shown to accurately reflect information and trends in other national databases (\eg transplant registry, comorbidity patterns)~\cite{kidneyehrcosmos, cosmospheno}. At the time of writing, most of the 121 published studies using Cosmos focus on describing the epidemiology or outcomes of specific conditions or services~\cite{nofzinger2025vaccine, chowdhury2025epidemiology, son2021coronavirus}.

The current work is distinct insofar as most of our objectives are predictive rather than descriptive or associative---we predict diagnoses, acute exacerbations of chronic disease, and utilization outcomes. With that in mind, some of the conditions and outcomes we explored in this evaluation have been studied at a smaller scale in previous research using Cosmos data. These studies highlight challenges and opportunities. \citet{cosmosascvd} used Cosmos to demonstrate that the types of subtle differences \modelname captures impact risk of acute events among patients with \ac{T2DM}. \citet{patel2024phenotype} demonstrated that Cosmos rates of \ac{SLE} based on structured data in diagnosis codes were lower than expected, and paradoxical mortality findings in \cite{hofmeister2024survival} highlight the challenges of working with \ac{EHR} data. Relatedly, Cosmos has been used for developing and validating risk scores and strata in other conditions, including cancer-associated thrombosis \cite{li2024external}, vision impairment in multiple sclerosis \cite{buxton2023vision}, and perinatal cardiovascular events among patients with rare congenital diseases \cite{wilkie2025maternal}. Several Cosmos publications have studied healthcare utilization outcomes in distinct subpopulations~ \cite{gottlieb2025delays, evans2025emergency, moin2025characteristics}. \autoref{sec:scaling_laws} demonstrates how forecasting these types of metrics can be scaled with \modelname. While the present evaluation did not include the detailed phenotype modeling and validation present in some of these studies, \modelname data can readily be analyzed and fine-tuned with greater specificity and logical complexity employed in other Cosmos studies. In this way, \modelname may inform and be informed by more traditional Cosmos research.

\subsection{Scaling laws for LLMs}
Research into the scaling behavior of machine learning models has fundamentally shaped the trajectory of \acp{LLM}. A pivotal study by \citet{scalinglawsopenai} showed that as we increase a model’s size, the amount of training data, and the compute budget, the performance improves following a smooth power-law trend---a predictable pattern that governs how model performance grows with more resources.
\citet{henighan2020scaling} extended this analysis beyond text to other domains like images, video, and multimodal tasks, finding that larger transformer models consistently yield better predictive power across modalities as compute and model size grow, again following power-law improvement curves. A further demonstration of scale came from \citet{gpt-3}, who presented 175B-parameter GPT-3 and found that simply making the model extremely large unlocked emergent capabilities; notably the ability to perform new tasks in a few-shot setting without any task-specific training. 
Subsequently, \citet{chinchilla} observed that many earlier large models were trained with too little data. They trained a 70B-parameter Chinchilla with four times more data than GPT-3, using the same compute budget, and found it outperformed much larger models.
The frontier of scale was later pushed further by \citet{chowdhery2022palmscalinglanguagemodeling} and \citet{gpt-4,gpt-4.5}, where performance continued to improve log-linearly with scale.
Studies of \ac{LLM} scaling laws provide a guiding principle for our work, which asks whether similar predictable gains occur when we scale up models on more patient data.

\subsection{Medical foundation models}
\label{sec:related_foundation}
Inspired by the success of foundation models in general domains, researchers have started developing large-scale models tailored to electronic health records.
BEHRT~\cite{li2020behrt} introduced a BERT-like transformer model trained on the longitudinal medical histories of 1.6 million patients.
This approach yielded substantial improvements for disease prediction compared to prior state-of-the-art models.
Med-BERT~\cite{rasmy2021med} was developed from 28 million patients, which further validated the power of large-scale pretraining on structured \ac{EHR} data.
CLMBR~\cite{stanford_clmbr} introduced an autoregressive next-code predictor trained on 3.4 million patient records; the learned representations improved AUCROC across multiple downstream tasks, especially in low-data settings.
MOTOR~\cite{stanford_motor} trained a transformer-based model for time-to-event (TTE) prediction using 55 million patient records, demonstrating the transfer learning ability for the TTE foundation model.

Moving beyond encoder-only models like BERT, recent research has explored generative and autoregressive transformers that can explicitly model the sequence of events in a patient’s timeline.
CEHR-GPT~\cite{cehr-gpt} was one of the first attempts to train GPT models on structured EHR data. It showed that the synthetic data generated by the model effectively captures the intricate patterns present in EHR data.
Introduced by \citet{ethos}, ETHOS is trained on tokenized event streams of patient health timelines and tasked with predicting the next set of events in a patient’s record.
Interestingly, it does so in a zero-shot fashion, effectively learning a simulator of possible futures for a patient.
In a similar vein, Event Stream GPT~\cite{2023eventstreamgptdata} provides tooling to convert complex, irregular medical event sequences into a format that a transformer can ingest, and methods to handle the causally ordered generation of events.
TransformEHR~\cite{yang2023transformehr} is an encoder–decoder transformer pretrained with a ``visit masking'' strategy: it masks out all the medical codes in some future visit and trains the model to reconstruct them from the preceding history.

Other work has focused on incorporating unstructured data into medical foundation models.
Foresight~\cite{foresight} integrated unstructured text with structured \ac{EHR} data, where important details from doctors’ notes were converted into standardized medical concepts and combined with coded data as input to a GPT-based model.
Foresight demonstrated the feasibility of training one large model to handle many prediction tasks across different institutions.

While these \ac{EHR}-focused foundation models have shown encouraging results, they also highlight challenges.
A recent comprehensive review by \citet{shakyllm} examined over 80 such models and found that many were trained on relatively narrow datasets (\eg, a single hospital’s records) and evaluated on surrogate tasks that may not translate to real clinical impact.
EHRSHOT~\cite{ehrshot} and FoMoH~\cite{fomoh} introduced new benchmark suites designed around patient timelines that extend beyond intensive care settings. These benchmarks emphasize robust, fair, and clinically meaningful evaluations. Instead of zero-shot approaches, both studies focused on smaller-scale models ($\sim\!100$M parameters) trained using a pretrain-then-finetune paradigm.
Recently, \citet{msr_scaling} conducted the first study on scaling laws for \ac{EHR} foundation models. They investigated how model performance scales with size and data volume at smaller scales using MIMIC-IV~\cite{mimic-iv}, identifying consistent patterns such as power-law relationships between compute resources, model parameters, and clinical utility.

\section{Methods}
\label{sec:methods}

\subsection{Data Pre-Processing}
\label{methods:data_preprocessing}

To ensure that Cosmos supplies the foundation model with records of sufficient depth and quality, we apply a three-stage filter before tokenization.

\paragraph{1. Patient selection.} We retain only adults who have meaningful longitudinal follow-up within Cosmos:
\begin{itemize}
\item Age between 18 and 120 years on January 1, 2012.
\item At least two successive face-to-face encounters within a two-year period between January 1, 2012 and April 17, 2025.
\end{itemize}
The encounter-frequency requirement screens out patients whose primary care occurs outside Cosmos-contributing sites, while the 2012 index date avoids the sparse documentation that characterizes earlier years and simplifies downstream temporal alignment. We exclude pediatric patients because of significantly different care patterns (\eg, routine well-child visits, growth-chart measurements, age-specific dosing).

\paragraph{2. Encounter selection.} From the eligible patients we include encounters based on these criteria:
\begin{itemize}
\item Encounter belongs to an eligible patient.
\item The encounter start date on or after January 1, 2012 and before April 17, 2025.
\item Encounter type is associated with direct clinical care (\eg, outpatient visit, emergency-department stay, inpatient admission, telehealth, and many others). We discard canceled appointments, test records, and administrative placeholders that rarely carry coded clinical data.
\end{itemize}

Encounter types span core clinical areas such as office visits (17.1\%), emergency (2.1\%), surgery (1.0\%), and inpatient (0.76\%), but also include a broad range of other encounter types from telemedicine (0.88\%) and home care (1.0\%) to anesthesia (0.97\%) and prenatal visits (0.39\%).

\paragraph{3. Post-filter cleanup.} Patients left with zero qualifying encounters after the above steps are removed to prevent empty timelines. This results in 118M unique patient records in our full dataset.

\paragraph{4. Train/Test Split.} 90\% of the patients were randomly allocated for training, with the remaining 10\% reserved for all evaluations. Experiments using temporal train/test splits to assess generalization across time are an important consideration for future work.

\vspace{0.75\baselineskip}

\noindent This pipeline yields a cohort whose records are dense enough to train a sequence model while minimizing noise from sporadic documentation and non-clinical artifacts. We applied minimal filtering of patient records or input data to reflect the diversity of patients and the realities of real-world healthcare documentation. For a full breakdown of our dataset, see \autoref{tab:supp_datset_statistics}.

\subsection{Tokenization Details}
\label{methods:tokenization}
Our tokenization method adapts a few key techniques from \citet{ethos}. In general, medical events are placed in chronological order according to the instant at which they were documented, with some type-specific nuances noted below. The token vocabulary is defined \textit{a priori} based on the individual events that are possible rather than frequency-based methods like byte-pair encoding. When possible we tokenized using ontologies with codes that directly capture hierarchical and categorical information, such as \ac{ICD-10-CM} and \ac{ATC}. All \modelname models have a vocabulary size of 7,105, which is summarized in \autoref{tab:curiosity_vocab}.

\begin{table}[!ht]
\centering
\small
\renewcommand{\arraystretch}{1.1}
\begin{tabular}{lccc}
\toprule
\textbf{Event Type} & \textbf{Tokens/event} & \textbf{Number of Unique Tokens}\\
\midrule
Sex & 1 & 6 \\
Race & 1 & 7 \\
Age and Years since 1970 & 1 & 24 \\
Beginning of Sequence & 1 & 1 \\
Encounter Starts \& Ends & 1 & 226 (113 types) \\
Department Specialties & 1 & 299 \\
Chief Complaint (Name) & 1 & 1231 \\
Chief Complaint (Body Location) & 1 & 67 \\
Diagnoses & 1-3 & 2429 \\
Lab Results & 1 & 1000 \\
Lab Quantiles & 1 & 10 \\
Medication Orders & 1-3 & 289 \\
Procedures & 1 & 1500 \\
Time & $\geq 1$ & 13  \\
\bottomrule
\end{tabular}
\caption{Medical events included in \modelname's vocabulary. The tokenizer also includes separation, padding, and unknown tokens.}
\label{tab:curiosity_vocab}
\end{table}

\subsubsection{Demographics}
Patient history sequences begin with a set of demographics tokens, which represent patient attributes that are not tied to a single encounter. We included demographics tokens for sex, race, and age at first encounter, along with the number of years from 1970 to the start of the patient's medical history (both in 5-year buckets). Additionally, we added a ``Beginning of Sequence'' token to denote the start of the patient timeline after the demographics section. Sex is bucketed into ``Male'', ``Female'', ``Unknown'', ``Masked'', ``Other'', and ``Unspecified''.

\subsubsection{Encounters}
Encounters are bookended with start- and end-encounter tokens---each with a denoted encounter type (\eg, ``Emergency\_Start'', ``Emergency\_End''). Each encounter header contains the start token, a department specialty token, and possible chief complaint tokens. Chief complaints in Cosmos consist of a name and an optional body location, which we separated into two unique consecutive tokens per chief complaint. When multiple encounters overlap, their encounter header and end-encounter tokens each appear at the proper instant in the timeline. In general, we do not enforce rules for overlapping or nested encounter-start or -end tokens; their placement in the patient sequence corresponds only to the time at which the event happened, with the tokens for a given encounter header always appearing consecutively.

\subsubsection{Diagnoses}
Diagnosis events are represented by the associated \ac{ICD-10-CM} code. \ac{ICD-10-CM} codes were split up into three tokens by category and sub-category (first 3 characters), specific details (characters 4-5), and additional details and extensions (characters 6-7). Diagnoses are represented in \modelname patient trajectories by 1-3 tokens depending on the specificity of the documentation in the patient's chart. Because diagnosis events in Cosmos only have documentation resolution at the day level, they are placed at the very beginning of the encounter after the encounter header or at midnight of their documented date, whichever comes later. Diagnoses with the same date are sequenced in random order.
\subsubsection{Labs}
\label{methods:tokenization:labs}
Our dataset contains the 1,000 most common numeric lab tests from Cosmos (representing \textgreater$99\%$ of all numeric lab test results). Each lab result is represented by two consecutive tokens which identify the test performed and the quantile of the numeric result, following ETHOS \cite{ethos}. Lab components are represented by Logical Observation Identifiers Names and Codes (LOINC\textsuperscript{\textregistered})\footnote{LOINC\textsuperscript{\textregistered} is a registered trademark of Regenstrief Institute, Inc.} codes, and numeric results are stratified by LOINC and measurement unit. For each LOINC-unit pair we partitioned values into 10 equal-frequency bins and mapped them to generic tokens. Compared to uniform-width binning, quantiles better accommodate non-linear mappings between heterogeneous units and better balance token frequencies, though they can compress rare but clinically meaningful extremes. The lab-result token pairs are placed in the sequence at the instant of collection. This ensures the sequence mirrors the patient's evolving clinical state but requires careful consideration to ensure that only information available at evaluation time is used.
\subsubsection{Medications}
Medication orders are represented by an \ac{ATC} code without other data. \ac{ATC} codes were split up into three sets of tokens by anatomical group and therapeutic subgroup (characters 1-3), pharmacological and chemical subgroup (characters 4-5), and chemical substance (characters 6-7). Medication orders are represented in \modelname patient trajectories using three consecutive tokens to represent the full \ac{ATC} code. They are placed in the sequence at the instant of the order.
\subsubsection{Procedures}
We extracted all billed procedures with Current Procedure Terminology (CPT\textsuperscript{\textregistered})\footnote{CPT\textsuperscript{\textregistered} is a registered trademark of the American Medical Association.} codes in Cosmos into our dataset. Our codes include such events as conventional procedures and imaging tests, as well as CPT codes associated with other billed codes like lab panel orders and level of service. We only tokenized the 1,500 most common procedures in our dataset to avoid sparse procedure tokens in our vocabulary (this represented 97.3\% of all procedure events). They are placed in the sequence at procedure start instant.
\subsubsection{Time}
The passage of time was represented in the medical event sequence by one or more tokens that represent a time interval. The token is selected from a set of time ranges following \citet{ethos}, ranging from ``1-5 minutes'' to ``3-6 months''. Events that occur within a shorter time span than the smallest time range are not separated using any time tokens. Events that occur more than 6 months apart have one or more ``6 month'' tokens, rounded to the nearest integer.
\subsubsection{Excluded Data}
\modelname is limited to structured data and does not include data such as clinical notes, images, or free-text results from diagnostic procedures. Additional structured data, such as vitals, allergies, and medication administrations, were not included at this time.

\subsection{Model Training Details}
\label{methods:training}

All \modelname models are built on the Qwen2 architecture. An overview of the architectural hyperparameters is shown in \autoref{tab:architectures}. All variants are trained with a context window of 8,192 tokens.

\begin{table}[!ht]
\centering
\small
\renewcommand{\arraystretch}{1.1}
\begin{tabular}{lccccc}
\toprule
\textbf{Model} & Params & Layers & Dimension & Heads & MLP dimension \\
\midrule
\modelname-S & 62M  & 6 & 768     & 12 & 3072 \\
\modelname-M & 119M & 12 & 768    & 12 & 3072 \\
\modelname-L & 1B   & 16 & 2048   & 32 & 8192 \\
\bottomrule
\end{tabular}
\caption{Summary of \modelname model configurations, including Small (S), Medium (M), and Large (L) model sizes. Abbreviation: MLP = multi-layer perceptron}
\label{tab:architectures}
\end{table}

\subsubsection{\modelname training and scaling laws}
\modelname models are built on the Qwen2 architecture \cite{qwen2}. Qwen2 incorporates pre-layer normalization, SwiGLU activations, rotary positional embeddings, and grouped-query attention, all of which improve training time and stability. \modelname models were trained with cross-entropy loss and standard hyperparameters, slightly adjusting them as determined by experimentation to account for possible differences in training transformers on medical events rather than natural language. Batch size was fixed at 512 sequences, and input sequences were densely packed in order to fully use the context window during training, with only a separation token marking different patients since our experiments demonstrated this was sufficient for training. Following \citet{chinchilla}, we employed a 10x learning rate decay with cosine schedule and used AdamW as our optimizer.

We estimated the compute $C$ used in training a model by the number of \acp{TFLOP} required in the forward and backward passes, using PaLM's methodology~\cite{chowdhery2022palmscalinglanguagemodeling}, where $N$ is the number of parameters in the model and $D$ is the number of training tokens used. To obtain our power-law fit for optimal parameter count $N_\text{opt}$ and number of tokens $D_\text{opt}$ for a fixed $C$, we fixed $C$ and varied $N$ and $D$ to obtain a set of isoFLOP experiments (\autoref{fig:scaling_laws}[A]). Each isoFLOP's loss-versus-$\log N$ and loss-versus-$\log D$ points were fit to parabolas, and the minimum point of the parabola served as $N_\text{opt}$ and $D_\text{opt}$ for that value of $C$. We then plotted $\log N_\text{opt}$ vs. $\log C$ (\autoref{fig:scaling_laws}[B]) and $\log D_\text{opt}$ vs. $\log C$ (\autoref{fig:scaling_laws}[C]) and fit a power law, using the first-degree $\alpha$ and $\beta$ terms, respectively, to measure the relative power-law scaling of optimal model size versus optimal training tokens.

\subsection{Benchmarking with Task-Specific Supervised Models}
\label{methods:supervised_baselines}

To contextualize the performance of \modelname, we implemented three baseline approaches representing distinct modeling paradigms. These included (1) linear and logistic regression models, representing classical linear approaches; (2) gradient-boosted decision trees, a strong non-linear method widely used for structured data; and (3) supervised transformers trained from a random initialization, a flexible deep learning architecture that retains the temporal information of each patient record. Each baseline model was trained independently for its corresponding downstream task and evaluated using the same datasets and procedures as those applied to the \modelname foundational models.

\paragraph{Sample construction.}
For each downstream task, we applied the evaluation's labeling logic to the \modelname training set to construct supervised examples. We used all supervised examples available in the \modelname training dataset up to a maximum of 5 million patient histories per task to fit the baselines. These samples were grouped by patient id and stratified by the task label where applicable. We reserved an additional 625{,}000 stratified rows when available or $10\%$ of the available training rows as a validation set for early stopping and hyper-parameter selection, leaving the designated \modelname development and test splits untouched for final evaluation.

\paragraph{Input representation}
With the exception of the supervised transformer baselines, input prompts were converted into a bag-of-words (BoW) count vector over the full \modelname vocabulary (\autoref{tab:curiosity_vocab}). Columns were normalized such that the maximum absolute value of each feature in the training set was 1.0 before fitting the linear and logistic models; XGBoost consumed raw counts. Preliminary experiments with inverse-frequency class-weighting showed no material AUCROC gain on these large samples, so we report unweighted results for simplicity.

\paragraph{Linear and logistic regression.}
We trained linear models with mini-batch stochastic gradient descent for up to 1,000 epochs over the five million row dataset with early stopping after 5 epochs without improvement. Classification used a logistic loss; regression used a squared loss with an L1 regularization penalty.

\paragraph{Gradient-boosted decision trees.}
We used a gradient-boosted decision tree classifier or regressor with \texttt{n\_estimators}$=10{,}000$, \texttt{max\_depth}$=6$, \texttt{learning\_rate}$=0.1$, and \texttt{subsample}$=0.8$. Training employed early stopping with a patience of $100$ boosting rounds. Other hyper-parameters followed XGBoost 1.5.2 library defaults unless specified.

\paragraph{Supervised Transformer.}
We trained the closest task-specific supervised counterpart to \modelname-M by using the same architecture as \modelname-M (\autoref{tab:architectures}) and the same tokenized input prompts. Unlike the bag-of-words inputs, this baseline preserves model architecture and the temporal sequence information in the original prompts. We attached a task-specific classification or regression head to the model and optimized with AdamW (weight decay 0.01), using a linear-decay schedule with warmup (500 warmup steps) and a peak learning rate of \(1\times10^{-4}\). Training ran for \(10{,}000\) update steps (global batch size \(128\); \(1.28\)M samples). We applied gradient-norm clipping at \(1.0\) and dropout 0.0.

We evaluated the supervised transformer baseline on a representative subset of tasks. The results are summarized in \autoref{fig:perf_diff_supervised_models} and \autoref{tab:los_baseline_results}. Across the evaluated tasks, XGBoost consistently outperformed logistic regression and was generally stronger than the transformer baseline, despite requiring less compute. Based on these findings, we used XGBoost as the primary task-specific supervised comparator but always report the strongest baseline result obtained. Estimating the value added by pretraining and the transfer-learning it enables would require a deeper analysis of performance saturation under an unbounded task-specific compute budget and is left to future work.

\subsection{Inference Details}
\label{methods:inference}
Inference with \modelname was performed by generating future medical event sequences via Monte Carlo sampling. For a given inference case, a patient history up to a specific moment was given as context to the model, and the model generated a specified number $d$ of output tokens $n$ times at a temperature of 1, where $d$ and $n$ depended on the needs of the individual task. If the amount of patient history exceeded the size of the model's context window (8,192 tokens for all \modelname models), the history is left-truncated as necessary. When this is necessary, the model loses the ability to see demographics and older information, but this aligns with how the model sees truncated sequences during pretraining.

All of our evaluations ask questions about events over a specified time range, not number of tokens. The model's $i$th generation is a series of tokens $(y_{i1}, y_{i2}, ..., y_{id})$, where $y_{ij} \in V$ are tokens in the vocabulary $V$. Times are assigned by incrementing the current time at generated time tokens based on the geometric midpoint of the interval. For each token $y \in V$ there is an associated time bucket $\left(t_{\text{min}}^{(y)}, t_{\text{max}}^{(y)}\right)$, which is (0, 0) for non-time tokens, and the geometric midpoint is given by $\Delta t_{y} = \sqrt{t_{\text{min}}^{(y)} t_{\text{max}}^{(y)}}$. Therefore, for generated tokens we create a sequence of event tuples $T_i=\big((y_{i1},t_{i1}), (y_{i2}, t_{i2}), ...,(y_{id},t_{id})\big)$, where $t_{ij} = t_{i(j-1)} + \Delta t_{y_{i(j-1)}}$ and $(y_{i0}, t_{i0})$ is the final event-time tuple in the model's context window. Because generated trajectories might not reach the full time length when generating by number of tokens, generated trajectories must be right-censored, and so the probability that a token in the target set $S$ occurs within time $\tau$ when using $n$ generations is given by:

\begin{equation}
P_{\tau}(S) = \frac{\sum_{i=1}^{n}{\mathds{1}[\exists j : y_{ij} \in S \text{ and } t_{ij} \leq \tau]}}{\sum_{i=1}^{n}{\max(\mathds{1}[t_{id} > \tau],\mathds{1}[\exists j : y_{ij} \in S \text{ and } t_{ij} \leq \tau]})}
\end{equation}

If the denominator is zero, we exclude the patient from the evaluation. Similar logic is used for getting the probability of counts of events in generated sequences, as well as the distribution of time-to-event outcomes. Future work is needed to improve model generations to always reach $\tau$ regardless of output length.

\subsection{Evaluation Details}
\label{methods:eval_details}
All evaluations were performed using patients from the held-out test set.

\subsubsection{Plausibility statistics}
\label{methods:plausibility}
We selected 20,000 random patients and performed generations starting from the end of the patient's last encounter that ended prior to 2022. To select for patients with some minimal degree of activity within health systems contributing data to Cosmos, we required that these patients have at least one prior encounter as well as one encounter beyond a year's time from the generation start. We used $n=25$ generations for each patient, using up to 2,000 tokens to reach one year's time, and discarding those generations that did not reach time. We used these generations to measure the percent of multi-token events that were invalid (\autoref{sec:validity}), the prevalence and co-occurrence rates (\autoref{sec:prevalence}), and the probability calibrations of the number and types of encounters (\autoref{sec:utilization_calibration}). For prevalence and co-occurrence, we measured the overall agreement between ground truth and predicted rates using \ac{RMSLE}:

\begin{equation}
    \text{RMSLE}=\sqrt{\sum_{i=1}^N {\left[ \log_{10}(x_i + \epsilon) - log_{10}(\hat{x}_i+\epsilon)\right]^2}}
\end{equation}

where $N$ is the number of event concepts, and $(x_i, \hat{x}_i)$ are the prevalence or co-occurrence rate of individual concepts in ground truth and generated sequences, making \ac{RMSLE} essentially the log-transformed \ac{RMSE}. For the probability calibration plots, we pooled patients into cohorts using equal quantile bin edges, and in each probability cohort measured the average predicted probability and the ground truth positive fraction. We quantified overall calibration for each count bucket using \acf{ECE}, which is just the \ac{MAE} of the calibration curve with respect to the diagonal (\ie, perfect calibration).

\subsubsection{Single-encounter generations}
\label{methods:encounter_generations}
Encounters were selected at random from across the test set without filtering for patients with a certain amount of previous history. This was done to mimic real clinical practice. We evaluated encounters that were either office visits, emergency visits, or inpatient admissions from the emergency department. For each encounter type, 10,000 encounters were selected, and all the medical events that occurred between encounter start and encounter end were tabulated, specifically for diagnosis, medication, lab, and procedure data types. Prompts terminated at the end of the encounter header (\ie, encounter type, department specialty, and any chief complaints), and generations stopped once the model generated the appropriate encounter end token. Twenty generations were used per sample, with up to 2,000 tokens allowed per generation. In order to be considered a true positive, the exact same code had to appear in both the patient simulation and the ground truth encounter, even for multi-token diagnosis and medication events. Micro-averaged precision and recall were computed for each encounter type and data type using different thresholds, and \ac{PR-AUC} was determined from these precision-recall curves.

For reference points, we pooled together the medical events from the patient's history prior to the encounter and measured the precision and recall of this method. We did this for different lookback windows, indicated by the gray dots in \autoref{fig:encounter_generation}.

\subsubsection{Disease-Specific Outcomes}
\label{methods:disease_outcomes}
Each sample provided to \modelname includes a patient's health timeline from the first token up to and including the order of a medication indicated for one of the following conditions: \ac{T2DM}, hypertension, and hyperlipidemia. When a sample exceeded the available context or left too little room for generation, we applied left truncation, removing tokens from the beginning of the sample. Given a sample, \modelname generated up to 2,000 new tokens from which predictions were made programmatically. 

Binary predictions, such as determining whether an adverse event occurred within a determined time $t$, were made by scanning the generation up to time $t$ for a set of \ac{ICD-10-CM} codes that describe an adverse outcome. Continuous predictions of lab values were made by uniformly sampling from within the range of values encapsulated by the relevant lab's quantile bucket. If multiple labs of the desired type were produced in a given patient simulation, the average of the uniform samples was used. For all prediction types, generations that did not reach the time threshold $t$ required by outcome definitions were discarded. \modelname was evaluated on 30,000 samples per condition and predictions extracted from the model's generations were micro-averaged over 80 generations per sample prior to calculating metrics.

\subsubsection{Incident Disease Risk Prediction}
\label{methods:population_screening}
All tasks here were formulated as binary prediction of a patient disease state within a 2-year window. Each prediction point in this cohort is unique to one patient and was chosen to be the last instant of a randomly selected outpatient encounter between 2020 and 2022. All patients were required to have at least 2 encounters of either outpatient or ``emergent'' encounter types in the 2 years prior to the prediction date. ``Emergent'' encounters are inpatient admission, emergency department, and urgent care visits. Additionally, all patients were required to be at least 18 and less than 120 years old at the prediction date to be included in these cohorts. For each incident disease prediction tasks, patients are excluded if they have had any of the diagnosis codes of any type from the phenotype definition before the prediction point. 

Each target's phenotype was classified as ``chronic'' or ``emergent'' and defined by a list of \ac{ICD-10-CM} codes within certain encounter types (see \autoref{table:pop_aoc_definitions}). A chronic target was marked positive if a diagnosis appeared in at least two outpatient encounters or in a single emergent encounter in the patient's history. An emergent target was marked positive only if a diagnosis occurred during an emergent encounter. Only encounter diagnoses and billing diagnoses were used for this target gathering. Patients in the ``Dementia (2yr)'' cohort were also required to be 60 years or older at the prediction date to make the task more difficult.

Roughly 5K patients and prediction points were sampled from the test set for each task (see \autoref{table:supp_eval_statistics} for exact counts). Since there were broad inclusion and minimal exclusion criteria, the dataset was naturally very imbalanced with all tasks having \textless$\!1.5\%$ positive prevalence. Random upsampling on the minority class was performed for each task cohort so at minimum 500 positive samples are present. During the calculation of performance metrics, samples were weighted by the inverse of the resampling factor so that the class proportions in the analysis matched those of in the full test set:

\begin{align}
w_{+} &= \frac{P^{\text{original}}_{+}}{P^{\text{resample}}_{+}},\\
w_{-} &= \frac{P^{\text{original}}_{-}}{P^{\text{resample}}_{-}},
\end{align}

where $w_{+}$ and $w_{-}$ are the weights assigned to positive and negative samples, and $P^{\text{original}}_{\pm}$ and $P^{\text{resample}}_{\pm}$ denote the number of positive/negative cases before and after up-sampling, respectively. These weights restore the population-level prevalence, ensuring that prevalence-dependent metrics remain unbiased despite the synthetic inflation of positives.

Inference on these tasks followed our typical strategy with the following modifications. We generated $n=60$ simulations for each patient initially with 2,000 tokens, and we retried up to two times to get the patient generations to the 2-year prediction time duration. When evaluating patient timelines generated by \modelname, the timeline would be evaluated to have the target if the timeline has any occurrence of any \ac{ICD-10-CM} codes from the target phenotype within the corresponding encounter types for that phenotype.

\subsubsection{Acute on Chronic Event Prediction }
\label{methods:acute_on_chronic}

The non-task specific inclusion criteria and the prediction date selection criteria used for incident disease risk prediction were also used as the base criteria for acute-on-chronic event prediction, with additional task-specific inclusion criteria applied for each condition. Each acute-on-chronic event prediction task consists of a chronic and an acute (``emergent'') phenotype, such as sickle cell disease and sickle cell crisis respectively. Find the full list of phenotypes in \autoref{table:pop_aoc_definitions}). 

To be included in the prediction cohort, a patient must meet both the base inclusion criteria and have the chronic phenotype before the prediction date. This means they must have at least two outpatient encounters or one emergent encounter with a clinical or billed diagnosis code from the phenotype before the prediction date.

To meet the acute phenotype criteria, a patient must have an emergent encounter with an encounter or billing diagnosis code within the prediction window in their ground truth data. Patients who have already had this acute event prior to the prediction date are still included within the cohort. A patient timeline generated by \modelname would be evaluated to have the target if the timeline has any occurrence of any \ac{ICD-10-CM} codes from the acute phenotype within an emergent encounter. The same sampling strategy, sample weighting for performance metrics, and sampling parameters as incident disease risk prediction are used in this evaluation. \autoref{table:supp_eval_statistics} has more details on dataset size and positive prevalence.

\subsubsection{Differential diagnosis}
\label{methods:differential_diagnosis}
For the \ac{HPB} and rheumatic diagnosis clusters, we selected 9 conditions from each to represent a range of prevalence and disease types. Specific definitions of each are in \autoref{table:diff_dx_definitions}. Patients were selected for evaluation based on the presence of relevant diagnosis codes in their record. We used retrospective inclusion based on the diagnosis outcomes in order to focus this evaluation on patients who have one of the target diagnoses. In order to be included in a disease cohort, the patient had to have at least two occurrences of one of the eligible \ac{ICD-10-CM} codes at separate encounters (this was done to mitigate diagnoses documented as part of rule-out diagnostic tests but for which the patient did not receive additional care). Time was measured relative to the encounter containing the first occurrence of an eligible \ac{ICD-10-CM} code, which was then used as the index time $t=0$. Additionally, to be included, patients were required to both have an encounter \textit{prior to} one year before the index diagnosis date as well as one encounter \textit{within} one year before the index diagnosis date. We then selected patients who received one of these diagnoses. Once a patient was selected, the first occurrence of each off-target diagnosis was identified for that patient. Selecting at most 9 random encounters from the year \textit{prior} to $t=0$, we had \modelname generate from the end of these encounters using $n=40$ generations and up to 2,000 output tokens to predict a patient's 18-month risk of having that diagnosis. Additionally, we also generated from the encounter header for the encounter at $t=0$ to measure \modelname's final predictions. We chose not to generate from within this encounter due to an increased chance of information leakage and to focus the evaluation on \textit{early} diagnosis detection and differential diagnosis.

For the training of the task-specific supervised models, we selected many additional patients. For a proper comparison to \modelname, we chose to include patients who were being worked up for, or were at risk of developing, a much wider set of diagnoses than \ac{HPB} or rheumatic diagnoses. This makes the comparison to \modelname more appropriate and also better reflects the diverse clinical scenarios a deployed system might encounter when flagging patients at risk of being diagnosed with specific diseases. For training, we expanded our pool of eligible patients by broadening the pool of possible inclusion diagnoses from those in our \ac{HPB} and rheumatic lists to all 3-character \ac{ICD-10-CM} codes, and selected multiple prediction dates for each as described above. We then trained 18 task-specific binary classifiers, one for each of the 18 \ac{HPB} and rheumatic diagnoses.

In order to visualize and contextualize the results, a patient's predicted risk over the subsequent 18 months was used as their diagnosis risk at a moment in time, and that risk remained the same until their next encounter for which we had model predictions. In \autoref{fig:hepato_diff_diagnosis} and \autoref{fig:rheum_diff_diagnosis}, the predictions are plotted for each day from 6 months out up to $t=0$. We first assessed \modelname's ability to flag at-risk patients, choosing a threshold probability of 10\%. Each individual plot only contains patients we had selected as having the indicated diagnosis at $t=0$. For each of the off-target lines in the plot, we also excluded patients whose ground truth record ever contained that off-target diagnosis (in the past or in the future), so that the off-target diagnosis lines represent the percent of patients flagged with an ``incorrect'' diagnosis.

\subsubsection{Forecasting patients' interactions with the health system}
\label{methods:utilization}
We used the same patients and \modelname generations described above in \autoref{methods:plausibility}. For each encounter type indicated, we computed the mean number of encounters predicted by \modelname in the following year, as well as that predicted by the task-specific regression model.

\subsubsection{Length of stay}
\label{methods:length_of_stay}
To evaluate \modelname's ability to predict inpatient length of stay, we randomly selected 10,000 Hospital Admissions from our test set and labeled each with the total length of stay from admission to discharge in seconds. For each hospital admission, we prompted \modelname with the patient's history up until the point of admission (including the department specialty and any associated chief complaint and flagged primary encounter diagnosis events). We then generated up to 2,000 new tokens with $n=20$ inpatient trajectories until the encounter's stop token appears. We took the median time-to-encounter-end as the model's prediction.

\subsubsection{30-day readmission}
\label{methods:readmission}
We constructed prediction targets by selecting 10,000 patients discharged between January 2020 and March 2025, subsequently determining whether these patients experienced a hospital readmission within 30 days. Readmissions occurring within 24 hours of discharge were excluded from this analysis because they may often reflect hospital transfers or clerically erroneous discharges \cite{cms2024HWR}. A 30-day readmission in a \modelname generation was defined as the beginning of a new inpatient encounter within 30 days of the discharge time.

\section{Acknowledgments}
We thank 
Zach Galvin and Zhuowen Nie for managing the high-performance computing environment; 
Andrea Noel, Kersten Bartelt, and Jackie Gerhart for carefully reviewing our evaluations to ensure clinical accuracy; 
Brian Olson and Amy Kim for creating and advising on the graphic designs; 
Samson Race Dorfman for his help reviewing code and data quality;
Phil Lindemann, Matthew Lungren, Jonathan Carlson, and Joe Petro for project guidance; 
the Epic Cosmos R\&D team for building and maintaining the essential infrastructure;
and the Cosmos Community for the courage demonstrated in creating Cosmos, the dataset that made this work possible.

\bibliographystyle{unsrtnat}
\bibliography{references}

\clearpage
\appendix
\titleformat{\section}{\normalfont\Large\bfseries}{Appendix \thesection:}{.5em}{}

\section{\modelname dataset statistics}
After data pre-processing as described in \autoref{methods:data_preprocessing}, data from Epic Cosmos was transformed into a subset used for the \modelname models, which was split into a train and test set at the patient level. Summary statistics of this dataset are shown in \autoref{tab:supp_datset_statistics}. Cosmos maintains and reports on over 1,000 metrics to assess data quality, focused on completeness, conformance, and plausibility. Some of these data quality metrics as measured on the \modelname dataset subset are displayed in \autoref{tab:supp_dataset_dq}.

\label{sec:dataset_stats}
\begin{table}[H]
  \centering
  \begin{tabular}{llrrr}
    \toprule
    \textbf{Characteristic} & \textbf{Group} &
    \textbf{Train} & \textbf{Test} &
    \textbf{Total} \\
    \midrule
    \multirow{10}{*}{Total Counts}
      & Patients         & 106 M    & 11.8 M    & 118 M \\
      & Events           & 104 B    & 11.5 B    & 115 B \\
      & Encounters       & 7.65 B   & 850 M     & 8.50 B \\
      & Diagnoses        & 15.3 B   & 1.70 B    & 17.0 B \\
      & Labs             & 15.8 B   & 1.76 B    & 17.6 B \\
      & Medications      & 7.35 B   & 817 M     & 8.17 B \\
      & Procedures       & 9.98 B   & 1.11 B    & 11.1 B \\
      & Tokens           & 136 B    & 15.1 B    & 151 B \\
      & Tokens/Patient   & 1278.3   & 1277.7    & 1278.3 \\
      & Tokens/Encounter & 17.7     & 17.7      & 17.7 \\
    \midrule
    \multirow{4}{*}{Age}
      & 18 - 39    & 43.6 M & 4.84 M & 48.4 M \\
      & 40 - 59    & 38.8 M & 4.31 M & 43.1 M \\
      & 60 - 79    & 21.2 M & 2.35 M & 23.5 M \\
      & 80+        & 2.90 M & 323 K  & 3.22 M \\
    \midrule
    \multirow{6}{*}{Race}
      & White                         & 74.9 M  & 8.33 M  & 83.3 M \\
      & BoAA     & 14.6 M  & 1.62 M  & 16.2 M \\
      & Asian                         & 4.67 M  & 519 K   & 5.19 M \\
      & AI/AN                         & 1.00 M  & 112 K   & 1.12 M \\
      & NHOPI                         & 386 K   & 42.7 K  & 429 K  \\
      & Other                         & 4.96 M  & 551 K   & 5.51 M \\
    \midrule
    \multirow{3}{*}{Ethnicity}
      & Not Hispanic or Latino        & 85.8 M & 9.54 M & 95.3 M \\
      & Hispanic or Latino            & 11.3 M & 1.26 M & 12.6 M \\
      & Unspecified                   & 9.34 M & 1.04 M & 10.4 M \\
    \midrule
    \multirow{6}{*}{Sex}
      & Female                        & 58.4 M & 6.49 M & 64.9 M \\
      & Male                          & 48.0 M & 5.33 M & 53.3 M \\
      & Unknown                       & 36.3 K & 4.10 K & 40.4 K \\
      & Masked                        & 4.97 K & 572    & 5.54 K \\
      & Other                         & 984    & 119    & 1.10 K \\
      & Unspecified                   & 877    & 91     & 968 \\
    \midrule
    \multirow{4}{*}{Number of source organizations}
      & 1                             & 61.9 M & 6.89 M & 68.8 M \\
      & 2                             & 31.0 M & 3.44 M & 34.5 M \\
      & 3                             & 10.1 M & 1.12 M & 11.2 M \\
      & 4+                            & 3.38 M & 376 K  & 3.76 M \\
    \bottomrule
  \end{tabular}
  \caption[\modelname dataset statistics]{\modelname dataset counts in the train, test, and full data sets, organized by various data types, demographics, and the number of source organizations contributing to each patient record. Abbreviations: BoAA = Black or African American, AI/AN = American Indian or Alaska Native, NHOPI = Native Hawaiian or Other Pacific Islander. Only breakdowns for First Race are shown.}
  \label{tab:supp_datset_statistics}
\end{table}

\begin{figure}[!ht]
    \centering
    \includegraphics[width=\linewidth]{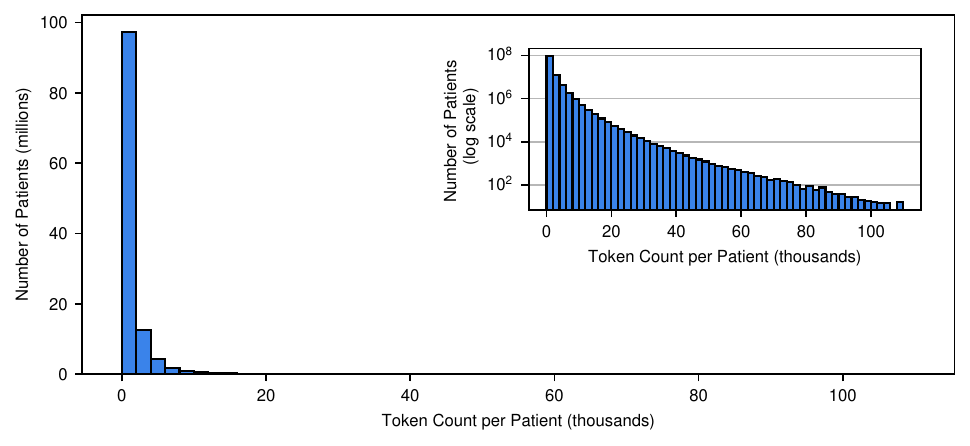}
    \caption[TokenDist]{\textbf{Distribution of token counts per patient.} Histogram of all patient token counts in the train and test sets, with a bin size of 2,000 tokens. Note: The inset plot has a log scaled y-axis to effectively visualize the long tail, and bins with counts less than or equal to 10 have been omitted.}
    \label{fig:TokenDist}
\end{figure}

\begin{table}[H]
  \centering
  \begin{tabular}{llr}
    \toprule
    \textbf{Data Type} & \textbf{Data Quality Metric} &
    \textbf{Percentage} \\
    \midrule
    \multirow{9}{*}{Patients}
      & Has first race           & 94.4\%  \\
      & Has ethnicity            & 91.2\%  \\
      & Has legal sex            & 100.0\% \\
      & Has birth date            & 100.0\% \\
      & Birth date after death    & 0.0\%   \\
      & Has diagnosis            & 98.4\%  \\
      & Has lab results          & 87.5\%  \\
      & Has medications          & 92.5\%  \\
      & Has problem list diagnosis             & 78.8\%  \\
    \midrule
    \multirow{8}{*}{Encounters}
      & Has start date                               & 100.0\% \\
      & Has end instant                              & 100.0\% \\
      & Has department specialty                     & 85.8\%  \\
      & Has chief complaint                          & 32.0\%  \\
      & Has admission instant (inpatient only)       & 100.0\% \\
      & Has discharge instant (inpatient only)       & 100.0\% \\
      & Has encounter type                           & 100.0\% \\
      & Start instant outside lifetime               & 0.1\%   \\
    \midrule
    \multirow{4}{*}{Medications}
      & Has medication code                          & 100.0\% \\
      & Has frequency                                & 88.2\%  \\
      & Has dose                                     & 79.4\%  \\
      & Min dose $<0$                                  & 0.0\%   \\
    \midrule
    \multirow{5}{*}{Labs}
      & Has LOINC code                               & 100.0\% \\
      & Has collected instant                        & 99.8\%  \\
      & Has resulted instant                         & 99.5\%  \\
      & Collected after resulted                     & 0.2\%   \\
      & Resulted $>30$ days after encounter end           & 0.2\%   \\
    \bottomrule
  \end{tabular}
  \caption[\modelname dataset data quality metrics]{\modelname dataset data quality metrics by data type}
  \label{tab:supp_dataset_dq}
\end{table}

\clearpage
\section{Tabulated results}
\label{sec:full_results}

\begin{table}[htbp]
\centering
\resizebox{\textwidth}{!}{
\begin{tabular}{
  >{\RaggedRight\arraybackslash}p{2.5cm}llrrr
}
\toprule
\textbf{Evaluation} & \textbf{Task} & \textbf{Metric} & \textbf{\modelname-S} & \textbf{\modelname-M} & \textbf{\modelname-L} \\

\midrule

\multirow{4}{=}{\raggedright Invalid multi-token events}
  & Diagnoses               & Percent & 0.057\% & 0.029\% & \textbf{0.011\%} \\
  & Medications             & Percent & 0.037\% & 0.023\% & \textbf{0.0087\%} \\
  & Lab results             & Percent & 0.0016\% & 0.0009\% & \textbf{0.0002\%} \\
  & Encounter headers       & Percent & 0.012\% & 0.011\% & \textbf{0.006\%} \\

\midrule

\multirow{4}{=}{\raggedright Event prevalence}
  & Diagnoses       & RMSLE & 0.311 & 0.292 & \textbf{0.281} \\
  & Labs            & RMSLE & 0.291 & 0.292 & \textbf{0.278} \\
  & Medications     & RMSLE & 0.240 & 0.233 & \textbf{0.215} \\
  & Procedures      & RMSLE & 0.311 & 0.289 & \textbf{0.264} \\

\midrule

\multirow{10}{=}{\raggedright Event co-occurrence}
  & Diagnosis-Diagnosis       & RMSLE & 0.117 & 0.107 & \textbf{0.102} \\
  & Diagnosis-Lab             & RMSLE & 0.178 & 0.174 & \textbf{0.170} \\
  & Diagnosis-Medication      & RMSLE & 0.104 & 0.101 & \textbf{0.097} \\
  & Diagnosis-Procedure       & RMSLE & 0.194 & 0.188 & \textbf{0.182} \\
  & Lab-Lab                   & RMSLE & 0.142 & 0.138 & \textbf{0.135} \\
  & Lab-Medication            & RMSLE & 0.118 & 0.118 & \textbf{0.116} \\
  & Lab-Procedure             & RMSLE & 0.216 & 0.215 & \textbf{0.210} \\
  & Medication-Medication     & RMSLE & 0.049 & 0.049 & \textbf{0.047} \\
  & Medication-Procedure      & RMSLE & 0.127 & 0.126 & \textbf{0.122} \\
  & Procedure-Procedure       & RMSLE & 0.170 & 0.168 & \textbf{0.164} \\

\midrule

\multirow{12}{=}{\raggedright Encounter frequency}
  & Office visit, 0             & ECE & 0.065 & 0.050 & \textbf{0.043} \\
  & Office visit, 1-2           & ECE & 0.058 & 0.057 & \textbf{0.051} \\
  & Office visit, 3-5           & ECE & 0.056 & \textbf{0.049} & 0.050 \\
  & Office visit, 6+            & ECE & 0.068 & 0.055 & \textbf{0.041} \\
  & Emergency, 0                & ECE & 0.035 & \textbf{0.028} & 0.030 \\
  & Emergency, 1                & ECE & 0.031 & 0.029 & \textbf{0.029} \\
  & Emergency, 2                & ECE & 0.028 & 0.025 & \textbf{0.022} \\
  & Emergency, 3+               & ECE & 0.024 & 0.020 & \textbf{0.018} \\
  & Inpatient, 0                & ECE & 0.027 & 0.022 & \textbf{0.017} \\
  & Inpatient, 1                & ECE & 0.029 & 0.027 & \textbf{0.024} \\
  & Inpatient, 2                & ECE & 0.011 & 0.011 & \textbf{0.009} \\
  & Inpatient, 3+               & ECE & 0.005 & 0.004 & \textbf{0.004} \\

\bottomrule
\end{tabular}
}
\caption{Percent, RMSLE, and ECE score comparisons across plausibility and encounter frequency tasks (\autoref{sec:aggregate_plausibility} and \autoref{sec:utilization_calibration}) for each \modelname model.}
\label{tab:model_comparison_plausibility}
\end{table}

\begin{table}[!htbp]
\centering
\resizebox{\textwidth}{!}{
\begin{tabular}{
  >{\RaggedRight\arraybackslash}p{3cm}lrrrr
}
\toprule
\textbf{Evaluation} & \textbf{Task} & \textbf{Reference} & \textbf{\modelname-S} & \textbf{\modelname-M} & \textbf{\modelname-L} \\

\midrule

\multirow{12}{=}{\raggedright Single-encounter generation}
  & Office visit, Diagnoses     & 0.098 & 0.480 & 0.502 & \textbf{0.548} \\
  & Office visit, Labs          & 0.129 & 0.312 & 0.331 & \textbf{0.421} \\
  & Office visit, Medications   & 0.075 & 0.176 & 0.195 & \textbf{0.251} \\
  & Office visit, Procedures    & 0.127 & 0.520 & 0.552 & \textbf{0.638} \\
  & Emergency, Diagnoses     & 0.106 & 0.390 & 0.412 & \textbf{0.450} \\
  & Emergency, Labs          & 0.415 & 0.782 & 0.808 & \textbf{0.840} \\
  & Emergency, Medications   & 0.170 & 0.475 & 0.497 & \textbf{0.536} \\
  & Emergency, Procedures    & 0.158 & 0.522 & 0.550 & \textbf{0.600} \\
  & Inpatient, Diagnoses     & 0.171 & 0.408 & 0.431 & \textbf{0.469} \\
  & Inpatient, Labs          & 0.595 & 0.846 & 0.871 & \textbf{0.899} \\
  & Inpatient, Medications   & 0.320 & 0.552 & 0.574 & \textbf{0.616} \\
  & Inpatient, Procedures    & 0.237 & 0.537 & 0.565 & \textbf{0.619} \\

\bottomrule
\end{tabular}
}
\caption{\ac{PR-AUC} score comparisons for single-encounter generation tasks (\autoref{sec:encounter_generations}) for each \modelname model and various baseline lookback windows.}
\label{tab:model_comparison_enc_gen}
\end{table}

\begin{table}[!htbp]
  \centering
  \resizebox{\linewidth}{!}{%
  \begin{tabular}{llrrrr}
    \toprule
    \textbf{Condition} & \textbf{Task (time horizon)} &
     \textbf{Supervised} &\textbf{\modelname-S} & \textbf{\modelname-M} & \textbf{\modelname-L} \\
    \midrule
    \multirow{10}{*}{Type 2 Diabetes}
      & ASCVD (1yr)                                  &  \textbf{0.878} &0.861 & 0.867 & 0.875 \\
      & ASCVD (3yr)                                  &  \textbf{0.907} &0.857 & 0.871 & 0.894 \\
      & CKD Prog.\ Stage 2 $\!\rightarrow\!$ 3 (1yr)   &  0.710 &0.720 & 0.736 & \textbf{0.762} \\
      & CKD Prog.\ Stage 2 $\!\rightarrow\!$ 3 (3yr)   &  0.766 &0.739 & 0.757 & \textbf{0.785} \\
      & CKD Prog.\ Stage 3 $\!\rightarrow\!$ 4\textsuperscript{+} (1yr) &  0.774 &0.748 & 0.773 & \textbf{0.796} \\
      & CKD Prog.\ Stage 3 $\!\rightarrow\!$ 4\textsuperscript{+} (3yr) &  0.799 &0.740 & 0.775 & \textbf{0.813} \\
      & Diabetic Neuropathy (1yr)                    &  0.885 &0.900 & 0.905 & \textbf{0.914} \\
      & Diabetic Neuropathy (3yr)                    &  0.906 &0.881 & 0.895 & \textbf{0.911} \\
      & Diabetic Retinopathy (1yr)                   &  0.899 &0.889 & 0.894 & \textbf{0.908} \\
      & Diabetic Retinopathy (3yr)                   &  \textbf{0.910} &0.859 & 0.876 & 0.902 \\
      & HgbA1c $<$ 7 (60-120 days)                   &  \textbf{0.764} &0.724 & 0.752 & 0.761 \\
      & HgbA1c $<$ 9 (60-120 days)                   &  0.687 &0.724 & 0.760 & \textbf{0.756} \\
      & HgbA1c $<$ 11 (60-120 days)                   &  0.710 &0.726 & 0.738 & \textbf{0.760} \\
      & HgbA1c $<$ 12 (60-120 days)                   &  0.730 &0.661 & 0.689 & \textbf{0.736} \\
    \midrule
    \multirow{10}{*}{Hypertension}
      & ASCVD (1yr)                                  &  0.854 &0.837 & 0.842 & \textbf{0.862} \\
      & ASCVD (3yr)                                  &  \textbf{0.893} &0.851 & 0.865 & 0.889 \\
      & CKD Prog.\ Stage 2 $\!\rightarrow\!$ 3 (1yr)   &  0.707 &0.719 & 0.724 & \textbf{0.771} \\
      & CKD Prog.\ Stage 2 $\!\rightarrow\!$ 3 (3yr)   &  0.772 &0.757 & 0.768 & \textbf{0.811} \\
      & CKD Prog.\ Stage 3 $\!\rightarrow\!$ 4\textsuperscript{+} (1yr) &  0.748 &0.738 & 0.751 & \textbf{0.788} \\
      & CKD Prog.\ Stage 3 $\!\rightarrow\!$ 4\textsuperscript{+} (3yr) &  0.813 &0.736 & 0.760 & \textbf{0.813} \\
      & Heart Attack (1yr)                           &  \textbf{0.828} &0.784 & 0.786 & 0.828 \\
      & Heart Attack (3yr)                           &  \textbf{0.870} &0.806 & 0.811 & 0.853 \\
      & Stroke (1yr)                                 &  \textbf{0.836} &0.773 & 0.780 & 0.812 \\
      & Stroke (3yr)                                 &  \textbf{0.867} &0.787 & 0.780 & 0.840 \\
    \midrule
    \multirow{8}{*}{Hyperlipidemia}
      & ASCVD (1yr)                                  &  \textbf{0.860} &0.830 & 0.834 & 0.853 \\
      & ASCVD (3yr)                                  &  \textbf{0.892} &0.845 & 0.859 & 0.883 \\
      & Heart Attack (1yr)                           &  \textbf{0.855} &0.772 & 0.781 & 0.809 \\
      & Heart Attack (3yr)                           &  \textbf{0.878} &0.804 & 0.814 & 0.842 \\
      & Chronic Heart Failure (1yr)                  &  \textbf{0.938} &0.916 & 0.919 & 0.925 \\
      & Chronic Heart Failure (3yr)                  &  \textbf{0.945} &0.898 & 0.908 & 0.920 \\
      & Stroke (1yr)                                 &  \textbf{0.855} &0.796 & 0.788 & 0.815 \\
      & Stroke (3yr)                                 &  \textbf{0.879} &0.799 & 0.808 & 0.837 \\
    \bottomrule
  \end{tabular}
  }
  \caption[All Disease-Specific Outcome figures]{\ac{AUCROC} on disease-specific outcome tasks (\autoref{sec:disease_outcomes}) for each \modelname model and the best-performing supervised task-specific model.}
  \label{tab:adverse_outcomes_restructured}
\end{table}

\begin{table}[!htbp]
  \centering
  {
  \footnotesize 
  \setlength{\tabcolsep}{5pt}
    \resizebox{\textwidth}{!}{
    \begin{tabular}{@{}>{\raggedright\arraybackslash}m{5cm}*{8}{S}@{}}
        \toprule
        \multirow{2}{*}{\bfseries Task (time horizon)} 
          & \multicolumn{2}{c}{\bfseries Supervised}
          & \multicolumn{2}{c}{\bfseries \modelname-S}
          & \multicolumn{2}{c}{\bfseries \modelname-M}
          & \multicolumn{2}{c}{\bfseries \modelname-L} \\
    \cmidrule(lr){2-3}\cmidrule(lr){4-5}\cmidrule(lr){6-7}\cmidrule(lr){8-9}
      & \multicolumn{1}{c}{AUCROC} & \multicolumn{1}{c}{PR-AUC}
      & \multicolumn{1}{c}{AUCROC} & \multicolumn{1}{c}{PR-AUC}
      & \multicolumn{1}{c}{AUCROC} & \multicolumn{1}{c}{PR-AUC}
      & \multicolumn{1}{c}{AUCROC} & \multicolumn{1}{c}{PR-AUC} \\
    \midrule
    \shortstack[l]{CHF Exacerbation (2yr) /\\ CHF}
      & 0.788 & 0.475
      & 0.767 & 0.405
      & 0.799 & 0.475
      & \textbf{0.827} & \textbf{0.511} \\
    
    \shortstack[l]{Alcohol Withdrawal Syndrome (2yr) /\\ Alcohol Use Disorder}
      & 0.823 & 0.434
      & 0.830 & 0.441
      & 0.839 & 0.445
      & \textbf{0.857} & \textbf{0.490} \\
    
    \shortstack[l]{Sickle Cell Crisis (2yr) /\\ Sickle Cell Disease}
      & 0.886 & 0.801
      & 0.884 & 0.790
      & 0.898 & 0.814
      & \textbf{0.913} & \textbf{0.844} \\
    
    \shortstack[l]{COPD Exacerbation (2yr) /\\ COPD}
      & 0.820 & 0.447
      & 0.812 & 0.397
      & 0.832 & 0.448
      & \textbf{0.847} & \textbf{0.479} \\
    
    \shortstack[l]{Asthma Exacerbation (2yr) /\\ Asthma}
      & \textbf{0.835} & 0.285
      & 0.798 & 0.230
      & 0.827 & 0.293
      & 0.832 & \textbf{0.300} \\

    \bottomrule
  \end{tabular}
  }
  }
  \caption{\ac{AUCROC} and \ac{PR-AUC} on acute-on-chronic tasks (\autoref{sec:acute_on_chronic}) for each \modelname model and the best-performing supervised task-specific model.}
  \label{tab:roc-prauc-aoc}
\end{table}

\begin{table}[!htbp]
  \centering
  {
  \footnotesize
  \setlength{\tabcolsep}{5pt}
  \resizebox{\textwidth}{!}{%
  \begin{tabular}{@{}>{\raggedright\arraybackslash}p{5cm}*{8}{S}@{}}
    \toprule
    \multirow{2}{*}{\bfseries Task (time horizon)}
      & \multicolumn{2}{c}{\bfseries Supervised}
      & \multicolumn{2}{c}{\bfseries \modelname-S}
      & \multicolumn{2}{c}{\bfseries \modelname-M}
      & \multicolumn{2}{c}{\bfseries \modelname-L} \\
    \cmidrule(lr){2-3}\cmidrule(lr){4-5}\cmidrule(lr){6-7}\cmidrule(lr){8-9}
      & \multicolumn{1}{c}{AUCROC} & \multicolumn{1}{c}{PR-AUC}
      & \multicolumn{1}{c}{AUCROC} & \multicolumn{1}{c}{PR-AUC}
      & \multicolumn{1}{c}{AUCROC} & \multicolumn{1}{c}{PR-AUC}
      & \multicolumn{1}{c}{AUCROC} & \multicolumn{1}{c}{PR-AUC} \\
    \midrule
    COPD (2yr)
      & 0.828 & 0.084
      & 0.807 & 0.048
      & 0.826 & 0.110
      & \textbf{0.839} & \textbf{0.114} \\

    CHF (2yr)
      & \textbf{0.894} & 0.125
      & 0.873 & 0.091
      & 0.883 & 0.143
      & 0.891 & \textbf{0.184} \\

    Dementia (2yr)
      & \textbf{0.933} & 0.106
      & 0.893 & 0.063
      & 0.908 & 0.144
      & 0.921 & \textbf{0.152} \\

    Asthma (2yr)
      & \textbf{0.795} & 0.058
      & 0.770 & 0.030
      & 0.770 & 0.083
      & 0.782 & \textbf{0.065} \\

    Alcohol Use Disorder (2yr)
      & \textbf{0.822} & 0.027
      & 0.753 & 0.015
      & 0.768 & 0.026
      & 0.801 & \textbf{0.072} \\

    Heart Attack (2yr)
      & \textbf{0.852} & 0.063
      & 0.802 & 0.054
      & 0.804 & 0.069
      & 0.801 & \textbf{0.092} \\
    \bottomrule
  \end{tabular}}%
  }
  \caption{\ac{AUCROC} and \ac{PR-AUC} on incident disease risk prediction (\autoref{sec:screening}) for each \modelname model and the best-performing supervised task-specific model.}
  \label{tab:roc-prauc-chronic}
\end{table}

\begin{table}[!htbp]
\centering
\resizebox{\textwidth}{!}{
\begin{tabular}{
  >{\RaggedRight\arraybackslash}p{3.0cm}lrrrr
}
\toprule
\textbf{Evaluation} & \textbf{Task} & \textbf{Supervised} & \textbf{\modelname-S} & \textbf{\modelname-M} & \textbf{\modelname-L} \\

\midrule

\multirow{9}{=}{\raggedright Hepatopancreato-biliary}
  & Acute Pancreatitis            & 0.512 & 0.526 & 0.557 & \textbf{0.594} \\
  & Chronic Pancreatitis          & 0.565 & 0.677 & 0.703 & \textbf{0.733} \\
  & Pancreatic Cancer             & 0.481 & 0.774 & 0.789 & \textbf{0.818} \\
  & Cholecystitis                 & 0.548 & 0.637 & 0.659 & \textbf{0.706} \\
  & Cholangitis                   & 0.603 & 0.686 & 0.706 & \textbf{0.744} \\
  & Liver Cancer                  & 0.576 & 0.787 & 0.797 & \textbf{0.821} \\
  & Chronic Viral Hepatitis       & 0.438 & 0.723 & 0.771 & \textbf{0.833} \\
  & Alcoholic Liver Disease       & 0.528 & 0.806 & 0.830 & \textbf{0.873} \\
  & Non-Alcoholic Steatohepatitis & 0.495 & 0.679 & 0.710 & \textbf{0.774} \\

\midrule

\multirow{9}{=}{\raggedright Rheumatic}
  & Osteoarthritis                   & 0.497 & 0.671 & 0.679 & \textbf{0.696} \\
  & Rheumatoid Arthritis              & 0.491 & 0.605 & 0.624 & \textbf{0.664} \\
  & Psoriatic Arthritis              & 0.488 & 0.740 & 0.750 & \textbf{0.791} \\
  & Polymyalgia Rheumatica           & 0.500 & 0.618 & 0.671 & \textbf{0.724} \\
  & Systemic Sclerosis               & 0.504 & 0.636 & 0.618 & \textbf{0.663} \\
  & Systemic Lupus Erythematosus     & 0.469 & 0.689 & 0.699 & \textbf{0.732} \\
  & Mixed Connective Tissue Disease  & 0.609 & 0.687 & 0.687 & \textbf{0.701} \\
  & Polymyositis/Dermatomyositis     & 0.540 & 0.644 & 0.647 & \textbf{0.687} \\
  & Fibromyalgia                     & 0.484 & 0.693 & 0.712 & \textbf{0.750} \\

\bottomrule
\end{tabular}
}
\caption{AUCROC for differential diagnosis prediction (\autoref{sec:diagnosis}) at $t=0$ (\ie, the start of the encounter of the target diagnosis event) for each \modelname model and the best-performing supervised task-specific model.}
\label{tab:ddx_results}
\end{table}

\clearpage

\begin{table}[!htbp]
    \centering
    \renewcommand{\arraystretch}{1.2}
    \begin{tabular}{lrrrr}
        \toprule
        \textbf{Task (time horizon)} &  \textbf{Supervised} &  \textbf{\modelname-S}  &\textbf{\modelname-M} &\textbf{\modelname-L} \\
        \midrule
        Number of inpatient visits (1yr)  &  0.187 &  0.091  &0.090 & \textbf{0.088}\\
        Number of emergency visits (1yr)  &  0.519 &  0.372  &0.366 &\textbf{0.364} \\
        Number of office visits  (1yr)   &  2.361 &  1.859  &1.777 &\textbf{1.712} \\
        \bottomrule
    \end{tabular}
    \caption{MAE for one-year encounter count forecasting (\autoref{sec:utilization}) for each \modelname model and the best-performing supervised task-specific model.}
    \label{tab:mae_encounter_counts}
\end{table}

\begin{table}[!htbp]
    \centering
    \renewcommand{\arraystretch}{1.2} %
    \begin{tabular}{lcccc}
        \toprule
        \textbf{Task}&
        \textbf{Supervised} &
        \textbf{\modelname-S} &
        \textbf{\modelname-M} &
        \textbf{\modelname-L} \\
        \midrule
        Admissions under 7 days& 2.032 & 1.346 & 1.279 & \textbf{1.238} \\
        Admissions under 14 days& 2.267 & 1.904 & 1.843 & \textbf{1.757} \\
        All admissions& 3.339 & 3.091 & 3.006 & \textbf{2.851} \\
        \bottomrule
    \end{tabular}
    \caption{MAE on length of stay tasks (\autoref{sec:los_readmissions}) measured in days for each \modelname model and the best-performing supervised task-specific model.}
    \label{tab:mae_buckets}
\end{table}

\begin{table}[!htbp]
    \centering
    \renewcommand{\arraystretch}{1.2}
    \begin{tabular}{lr}
        \toprule
        \textbf{Model} & \textbf{AUCROC} \\
        \midrule
        \modelname-L   & \textbf{0.770} \\
        \modelname-M   & 0.723 \\
        \modelname-S   & 0.706 \\
        Supervised & 0.718 \\
        \bottomrule
    \end{tabular}
    \caption{\ac{AUCROC} scores on the 30-day readmission task (\autoref{sec:los_readmissions}) for the \modelname models and the best-performing task-specific model.}
    \label{tab:auc_scores}
\end{table}

\begin{table}[!htbp]
\centering
\resizebox{\textwidth}{!}{
\begin{tabular}{
  >{\RaggedRight\arraybackslash}p{3.0cm}lrrr
}
\toprule
\textbf{Evaluation} & \textbf{Task} & \textbf{Transformer} & \textbf{XGBoost} & \textbf{Linear Regression}\\

\midrule

\multirow{3}{=}{\raggedright Length of Stay}
  & All Admissions             & \textbf{3.34} & 3.57 & 3.67 \\
  & Admissions under 14 days    & \textbf{2.27} & 2.53 & 2.58 \\
  & Admissions under 7 days     & \textbf{2.03} & 2.34 & 2.35 \\
\bottomrule
\end{tabular}
}
\caption{MAE (days) on the length of stay task (\autoref{sec:los_readmissions}) across the task-specific supervised models, including a supervised 119M parameter transformer described in \autoref{methods:supervised_baselines}.}
\label{tab:los_baseline_results}
\end{table}

\clearpage

\section{Multi-token event validity}
\label{sec:validity}
Each of the \modelname models was evaluated for the frequency of generating invalid multi-token events. A diagnostic event is considered invalid if there are \ac{ICD-10-CM} tokens in combination that do not correspond to real \ac{ICD-10-CM} codes. The same was done for medication orders with \ac{ATC} codes. A lab result is considered invalid if the lab token is not followed by a lab result quantile token. Encounter headers were considered invalid if an encounter start token was not directly followed by a department specialty token (as noted in \autoref{tab:supp_dataset_dq}, not all encounters have a specified department specialty, but the encounter headers still have a token with a specialty type of 'unspecified'). These experiments were done with 20,000 patients, each with 25 1-year generations. The denominator for each metric is the number of tokens that initiated an event. 

\begin{figure}[!ht]
    \centering
    \includegraphics[width=\linewidth]{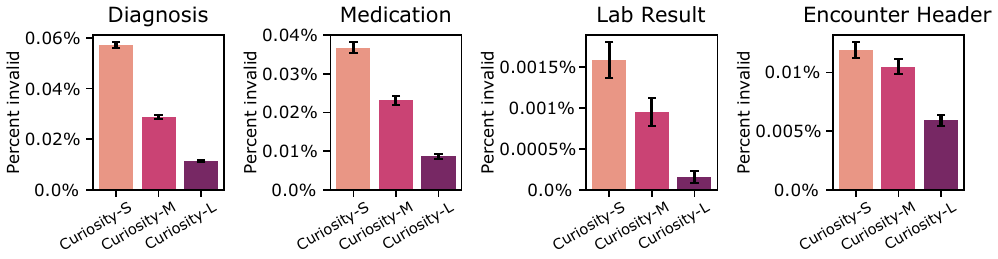}
    \caption[Validity]{\textbf{Generation syntactic validity.} The percent of diagnosis, medication, lab result, and encounter header events that were invalidly generated by each of the \modelname models.}
    \label{fig:supp_validity}
\end{figure}

\section{Event prevalence and co-occurrence}
\label{sec:prevalence}
\begin{figure}[!ht]
    \centering
    \includegraphics[width=\linewidth]{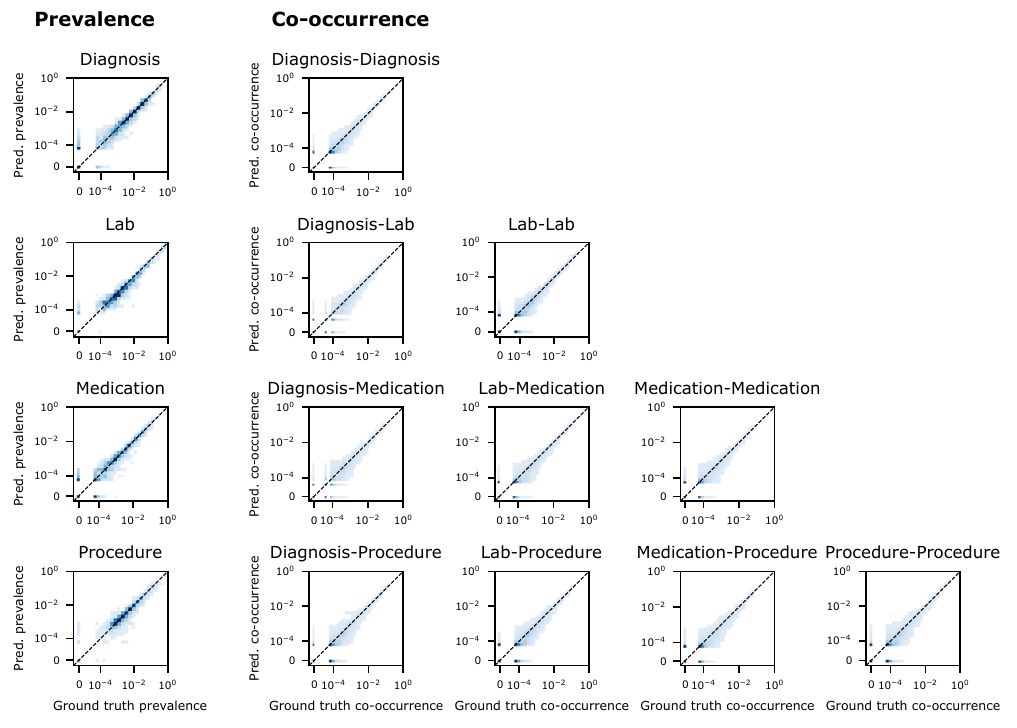}
    \caption[Prevalence]{\textbf{Plausibility of predicted medical event prevalence and pairwise co-occurrence.} On the left are heatmap scatter plots for the prevalence of different discrete medical events in \modelname-L's predicted generations versus the ground truth. The plots include diagnoses (first 3 \ac{ICD-10-CM} characters only), medications, labs, and procedures. On the right are heatmap scatter plots showing the fraction of patients in which pairs of medical events co-occurred within one year, comparing \modelname-L's predicted generations versus ground truth. Note that because these are all log-log plots, zero prevalence and co-occurrence values are also shown but with an explicit gap.}
    \label{fig:supp_prevalence}
\end{figure}

One way to assess the plausibility of \modelname's generations is to measure the prevalence of generated medical events. Given the model has to learn how to properly generate syntactically valid medical event sequences, it is not necessarily trivial to also learn the frequency at which different events of different types should appear for individual patients. For 20,000 patients, we generated 1 year's worth of sequences 25 times and measured, on average, how many times different medical events occurred. We measured this for diagnoses (at the 3-character \ac{ICD-10-CM} code level for simplicity), labs, medications, and procedures, as seen in \autoref{fig:supp_prevalence}.

In addition to measuring agreement in the prevalence of single events in generated sequences, we next asked whether pairs of events co-occur within a patient's generated sequence the proper amount. This second-degree question is a quick, aggregate measure that the model understands the relationship between events, both within and across data types. We measured the fraction of patients in which each possible pair of events co-occurred. Results in \autoref{fig:supp_prevalence} and \autoref{tab:model_comparison_plausibility} show that \modelname generally does well at generating events at the same rate of co-occurrence as observed in real patient medical event sequences.

\clearpage

\section{\autoref{sec:perf_v_loss} results}
\begin{figure}[!ht]
    \centering
    \includegraphics[width=\linewidth]{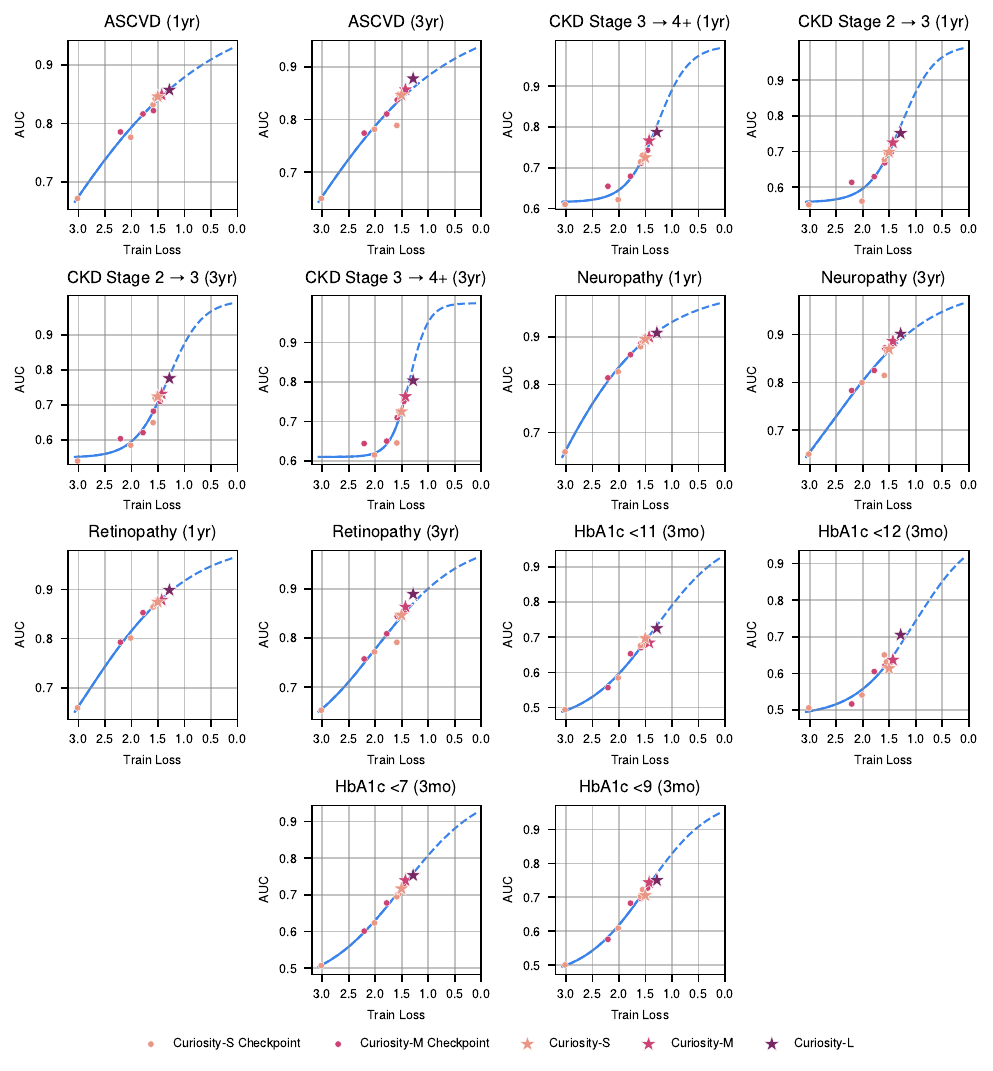}
    \caption[T2D performance vs loss]{\textbf{T2DM-specific outcome performance measured by AUCROC improves as train loss decreases.} We evaluated each \modelname model, along with earlier checkpoints from the \modelname-S and \modelname-M training runs, on all T2DM tasks. We fit a sigmoid curve to all points except for \modelname-L to assess the sigmoid curve’s predictive utility. We evaluated each model using a more conservative $n = 20$ simulations. Stars indicate compute-optimal models.}
    \label{fig:t2d_perf_v_loss}
\end{figure}

\begin{figure}[!ht]
    \centering
    \includegraphics[width=\linewidth]{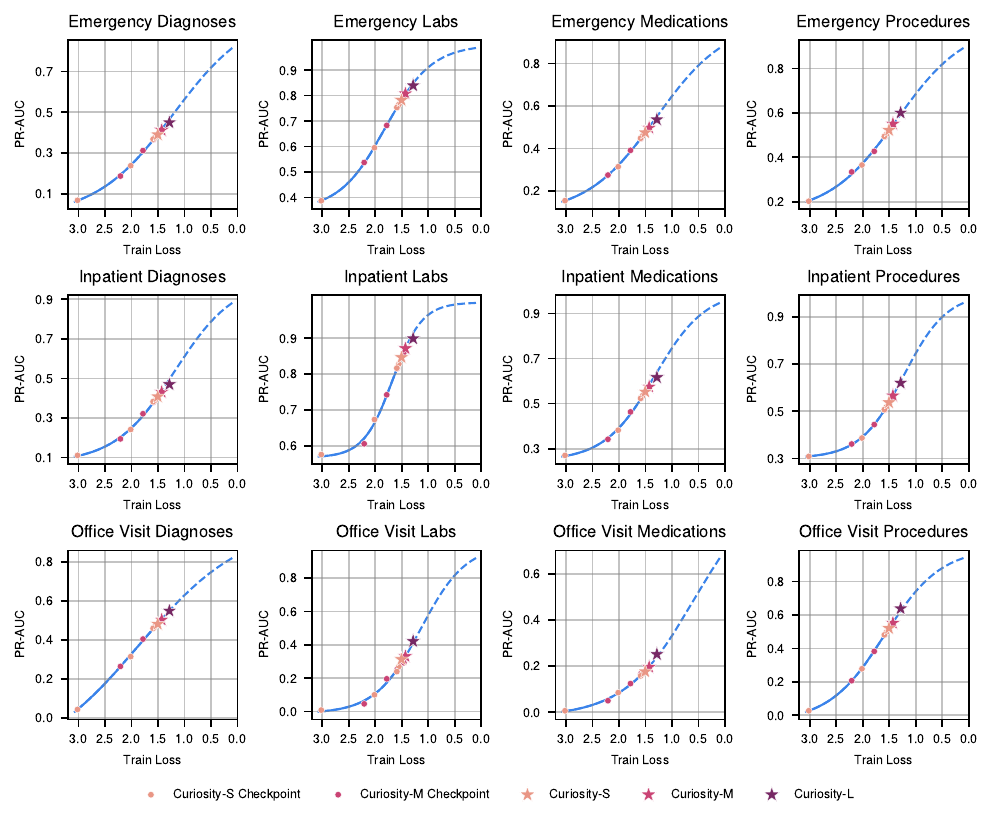}
    \caption[Encounter completions performance vs loss]{\textbf{Single-encounter generations improve as train loss decreases.} We evaluated each \modelname model, along with earlier checkpoints from the \modelname-S and \modelname-M training runs, on all of the single-encounter completion tasks. We fit a sigmoid curve to all points except for \modelname-L to assess the sigmoid curve’s predictive utility. We evaluated each model using $n = 20$ simulations. Stars indicate compute-optimal models.}
    \label{fig:el_perf_v_loss}
\end{figure}

\clearpage
\section{Bias and Fairness}
\label{sec:bias_fairness}

\begin{figure}[!ht]
    \centering
    \includegraphics[width=1\linewidth]{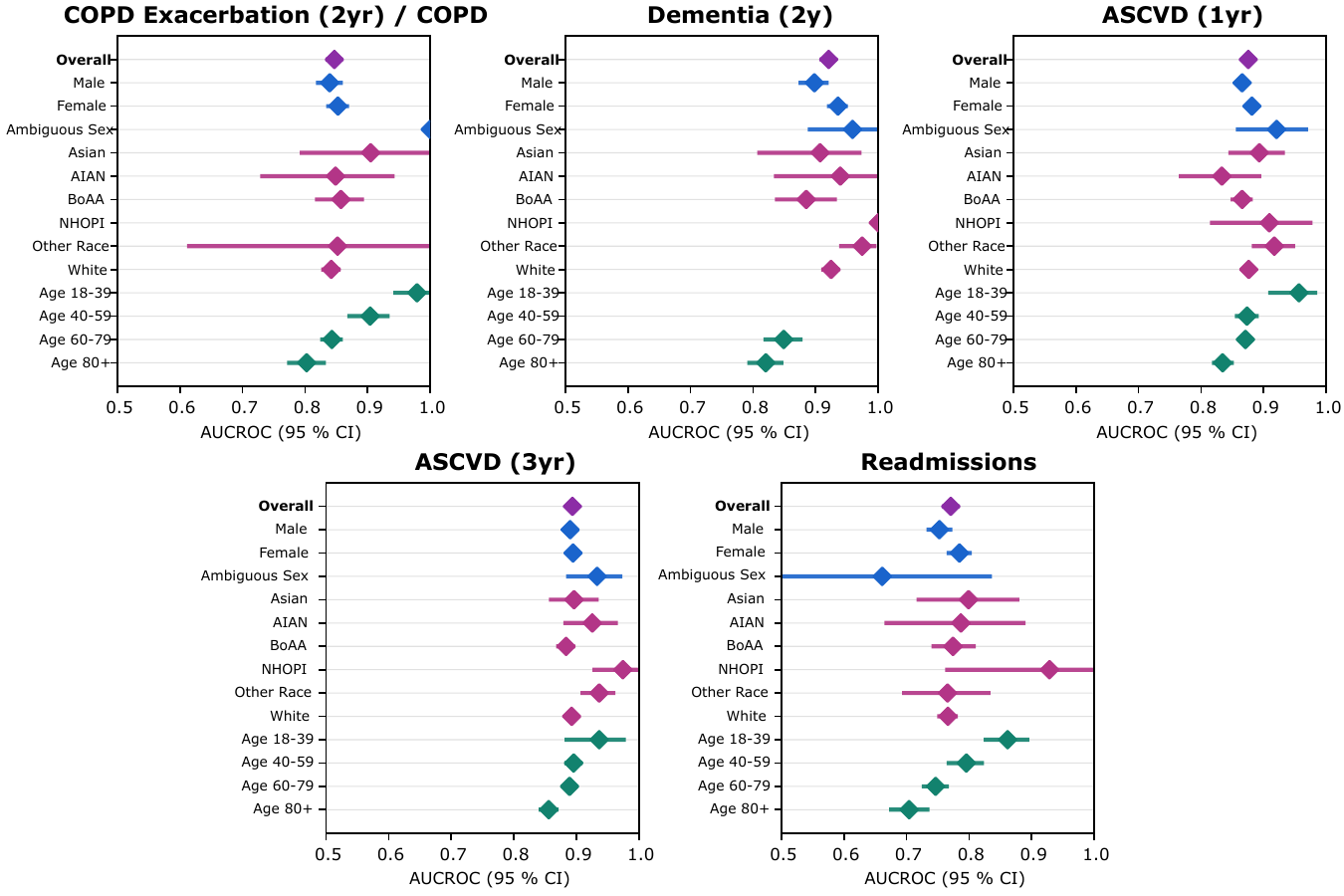}
    \caption[Bias and Fairness]{\textbf{Bias Impact via Subcohort Analysis.} \modelname-L's AUCROCs with 95\% bootstrapped CI stratified by demographic groups are displayed for five different predictive tasks for brevity. In order of which they appear they are: one from acute-on-chronic event, one from incident disease risk, two from \ac{T2DM}-specific outcomes, and one from operational outcomes.}
    \label{fig:supp_bias_and_fairness}
\end{figure}

\clearpage

\section{Additional Results for Disease-Specific Outcomes}
\label{appendix:more_outcomes}
\modelname predicts lab values that exhibit moderate correlation with actual observed values. \modelname predictions are noisy as a consequence of the limited expressiveness of the model's lab result value bucketed tokens (see \autoref{methods:tokenization:labs} for how labs are tokenized).

\begin{figure}[H]
    \centering
    \includegraphics[width=0.9\linewidth]{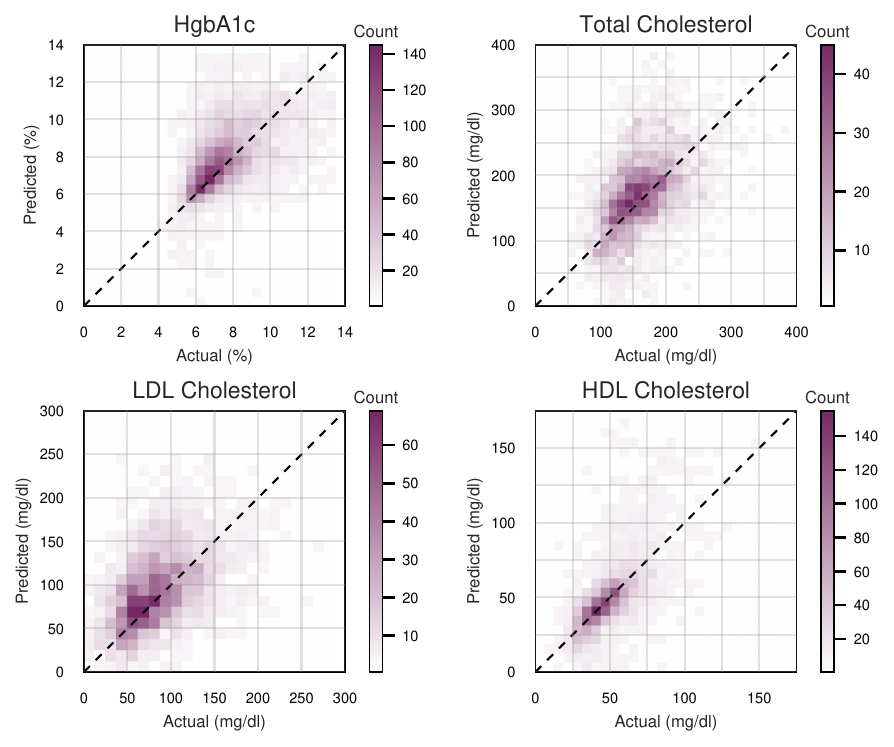}
    \caption[All Disease-Specific Outcome figures]{Heat map scatter plots comparing 3-month \modelname-L predictions of lab values to the patient's actual results. Plots show \ac{HgbA1c} for patients with diabetes and show total cholesterol, LDL cholesterol, and HDL cholesterol for patients with hyperlipidemia.}
    \label{fig:dso_lab_plots}
\end{figure}

\begin{table}[!ht]
\centering
\small
\begin{tabular}{lcccccccc}
\toprule
\textbf{Model} & \multicolumn{2}{c}{\textbf{HgbA1c}} & \multicolumn{2}{c}
{\textbf{HDL Cholesterol}} & \multicolumn{2}{c}{\textbf{LDL Cholesterol}} & \multicolumn{2}{c}{\textbf{Total Cholesterol}} \\
\midrule
& \textbf{MAE} & \textbf{RMSE} & \textbf{MAE} & \textbf{RMSE} & \textbf{MAE} & \textbf{RMSE} & \textbf{MAE} & \textbf{RMSE}\\
\midrule
\modelname-L & 1.21 & 1.69 & 11.57 & 18.30 & 31.51 & 41.91 & 43.01 & 58.66 \\
\modelname-M & 1.24 & 1.73 & 13.15 & 20.36 & 33.41 & 44.75 & 45.75 & 61.98\\
\modelname-S & 1.30 & 1.81 & 13.62 & 21.06 & 33.86 & 44.70 & 44.88 & 60.61\\
Supervised & \textbf{1.05} & \textbf{1.43} & \textbf{10.07} & \textbf{13.49} & \textbf{26.58} & \textbf{34.43} & \textbf{31.44} & \textbf{42.06}\\
\bottomrule
\end{tabular}
\caption{MAE and RMSE values for the corresponding subplots in \autoref{fig:dso_lab_plots} across all \modelname models and the best performing task-specific model.}
\label{table:dso_lab_metrics}
\end{table}

\clearpage
\section{Evaluation Dataset Characteristics}
\label{appendix:eval_statistics}

\begin{table}[!ht]
  \centering
  \resizebox{\linewidth}{!}{%
  \begin{tabular}{llrr}
    \toprule
    \textbf{Condition} & \textbf{Task (time horizon)} &
     \textbf{Test Sample Size} &\textbf{Positive Prevalence} \\
    \midrule
    \multirow{10}{*}{Type 2 Diabetes}
      & ASCVD (1yr)                                  &  20{,}653 & 0.137 \\
      & ASCVD (3yr)                                  &  12{,}331 & 0.353  \\
      & CKD Prog.\ Stage 2 $\!\rightarrow\!$ 3 (1yr)   & 1{,}462 & 0.159 \\
      & CKD Prog.\ Stage 2 $\!\rightarrow\!$ 3 (3yr) & 923 & 0.462 \\
      & CKD Prog.\ Stage 3 $\!\rightarrow\!$ 4\textsuperscript{+} (1yr) &  3{,}795 & 0.107 \\
      & CKD Prog.\ Stage 3 $\!\rightarrow\!$ 4\textsuperscript{+} (3yr) &  2{,}071 & 0.338 \\
      & Diabetic Neuropathy (1yr)                    &  21{,}879 & 0.337 \\
      & Diabetic Neuropathy (3yr)                    &  15{,}196 & 0.604 \\
      & Diabetic Retinopathy (1yr)                   &  20{,}272 & 0.110 \\
      & Diabetic Retinopathy (3yr)                   &  11{,}293 & 0.266 \\
      & HgbA1c $<$ 7 (60-120 days)                   &  5{,}608 & 0.401  \\
      & HgbA1c $<$ 9 (60-120 days)                   &  5{,}608 & 0.827 \\
      & HgbA1c $<$ 11 (60-120 days)                   &  5{,}608 & 0.950 \\
      & HgbA1c $<$ 12 (60-120 days)                   &  5{,}608 & 0.975 \\
    \midrule
    \multirow{10}{*}{Hypertension}
      & ASCVD (1yr)                                  &  21{,}344 & 0.180 \\
      & ASCVD (3yr)                                  &  13{,}687 & 0.419 \\
      & CKD Prog.\ Stage 2 $\!\rightarrow\!$ 3 (1yr) & 1{,}429 & 0.227 \\
      & CKD Prog.\ Stage 2 $\!\rightarrow\!$ 3 (3yr) & 985 & 0.541 \\
      & CKD Prog.\ Stage 3 $\!\rightarrow\!$ 4\textsuperscript{+} (1yr) &  4{,}157 & 0.120 \\
      & CKD Prog.\ Stage 3 $\!\rightarrow\!$ 4\textsuperscript{+} (3yr) &  2{,}326 & 0.354 \\
      & Heart Attack (1yr)                           &  20{,}301 & 0.049 \\
      & Heart Attack (3yr)                           &  11{,}139 & 0.158 \\
      & Stroke (1yr)                                 &  20{,}254 & 0.045 \\
      & Stroke (3yr)                                 &  10{,}939 & 0.138 \\
    \midrule
    \multirow{8}{*}{Hyperlipidemia}
      & ASCVD (1yr)                                  &  22{,}300 & 0.140 \\
      & ASCVD (3yr)                                  &  13{,}996 & 0.352 \\
      & Heart Attack (1yr)                           &  21{,}515 & 0.037 \\
      & Heart Attack (3yr)                           &  11{,}884 & 0.119 \\
      & Chronic Heart Failure (1yr)                  &  22{,}845 & 0.200 \\
      & Chronic Heart Failure (3yr)                  &  14{,}771 & 0.394 \\
      & Stroke (1yr)                                 &  21{,}540 & 0.036 \\
      & Stroke (3yr)                                 &  11{,}825 & 0.109 \\
    \bottomrule
  \end{tabular}
  }
\caption[Test set cohort composition]{Test-set sample sizes and positive class prevalence for each disease-specific adverse outcome (\autoref{sec:disease_outcomes}), grouped by clinical condition.}
\label{tab:dso_target_characteristics}
\end{table}

\begin{table}[H]
  \centering
  \small
  \begin{tabular}{lrr}
    \toprule
    \textbf{Task} & \textbf{Test Sample Size} & \textbf{Positive Prevalence} \\
    \midrule
    30-day Readmission & 10,000 & 0.1074 \\
    \bottomrule
  \end{tabular}
  \caption[30-day readmission cohort]{Test set sample size and positive class prevalence for the 30-day readmission task (\autoref{sec:los_readmissions}).}
  \label{tab:readmission_30d_stats}
\end{table}

\begin{table}[htbp]
\centering
\small
\begin{tabular}{ p{3cm} m{5.5cm} m{2cm} m{2cm} m{2cm} }
\toprule
\textbf{Task Type} & \textbf{Task (time horizon)} & \textbf{Patient Count} & \textbf{Resampled Positive Prevalence} & \textbf{Real \newline Positive Prevalence} \\
\midrule 
\multirow{4}{*}{Acute-on-Chronic}
    & Alcohol Withdrawal Syndrome (2yr) / \newline Alcohol Use Disorder & 5032 & 0.111 & 0.105 \\
    & Asthma Exacerbation (2yr) / \newline Asthma & 5199 & 0.097 & 0.039 \\
    & CHF Exacerbation (2yr) / \newline CHF & 5000 & 0.184 & 0.190 \\
    & COPD Exacerbation (2yr) / \newline  COPD & 5004 & 0.147 & 0.129 \\
    & Sickle Cell Crisis (2yr) / \newline Sickle Cell Disease & 1933 & 0.339 & 0.339 \\

\midrule
\multirow{9}{*}{Incident Disease Risk}
    & Alcohol Use Disorder (2yr) & 5369 & 0.093 & 0.003 \\
    & Asthma (2yr) & 5116 & 0.098 & 0.008 \\
    & CHF (2yr) & 5202 & 0.096 & 0.013 \\
    & COPD (2yr) & 5083 & 0.099 & 0.011 \\
    & Dementia (2yr) & 5369 & 0.093 & 0.006 \\
    & Heart Attack (2yr) & 5330 & 0.094 & 0.010 \\
\bottomrule
\end{tabular}
\caption{Summary statistics for acute-on-chronic (\autoref{sec:acute_on_chronic}) and incident disease risk (\autoref{sec:screening}). Note the sickle cell crisis task has fewer than 5,000 patients since there were only 1,933 patients who met inclusion criteria in the test set.}
\label{table:supp_eval_statistics}
\end{table}

\begin{table}[h!]
\centering
\begin{tabular}{lr}
\hline
\textbf{Diagnosis} & \textbf{Test Sample Size} \\

\midrule

Acute Pancreatitis              & 1239 \\
Chronic Pancreatitis            & 1002 \\
Pancreatic Cancer               & 980 \\
Cholecystitis                   & 1118 \\
Cholangitis                     & 967 \\
Liver Cancer                    & 970 \\
Chronic Viral Hepatitis         & 992 \\
Alcoholic Liver Disease         & 1027 \\
Non-Alcoholic Steatohepatitis   & 979 \\

\midrule

Osteoarthritis                  & 1256 \\
Rheumatoid Arthritis            & 1110 \\
Psoriatic Arthritis             & 951 \\
Polymyalgia Rheumatica          & 937 \\
Systemic Sclerosis              & 723 \\
Systemic Lupus Erythematosus    & 1015 \\
Mixed Connective Tissue Disease & 572 \\
Polymyositis/Dermatomyositis    & 466 \\
Fibromyalgia                    & 1067 \\

\hline
\end{tabular}
\caption{Summary statistics for the \ac{HPB} and rheumatic differential diagnosis evaluation sets (\autoref{sec:diagnosis}).}
\label{tab:ddx_stats}
\end{table}

\begin{table}[h!]
\centering
\begin{tabular}{l r r r}
\hline
\textbf{Category} & \textbf{N} & \textbf{Actual Standard Deviation} & \textbf{Actual Mean} \\
\hline
all admissions   & 10000 & 8.20  & 5.65   \\
admissions $\leq$ 4d  & 5503  & 0.95 & 2.28 \\
admissions $\leq$ 7d  & 7762  & 1.64  & 3.15  \\
admissions $\leq$ 10d & 8734  & 2.26 & 3.72  \\
admissions $\leq$ 14d & 9305  & 2.93  & 4.21  \\
admissions $\leq$ 50d & 9968  & 5.55 & 5.36  \\
\hline
\end{tabular}
\caption{Summary statistics for the length of stay evaluation set (\autoref{sec:los_readmissions}).}
\label{tab:los_stats}
\end{table}

\clearpage
\section{Disease Phenotypes}
\label{appendix:phenotypes}
Disease phenotypes for each category of evaluation task are built with a variety of criteria, such as encounter type inclusions, number of occurrences, and code groupers, which are described in \autoref{methods:eval_details}. Below are the code sets we used for defining each phenotype.

\begin{table}[htbp]
\centering
\begin{tabular}{ m{4cm} m{6cm} m{3.3cm} }
\toprule
\textbf{Disease Type} & \textbf{Disease} & \textbf{\ac{ICD-10-CM} Codes} \\
\midrule 
\multirow{9}{*}{Hepatopancreatobiliary}
    & Acute Pancreatitis & K85 \\
    & Chronic Pancreatitis & K86.0, K86.1 \\
    & Pancreatic Cancer & C25 \\
    & Acute Cholecystitis & K81 \\
    & Cholangitis & K83.0 \\
    & Chronic Viral Hepatitis & B18 \\
    & Liver Cancer & C22 \\
    & Alcoholic Liver Disease & K70 \\
    & Non-Alcoholic Steatohepatitis & K75.81 \\
\midrule
\multirow{9}{*}{Rheumatic}
    & Osteoarthritis & M15-19 \\
    & Rheumatoid Arthritis & M05, M06 \\
    & Psoriatic Arthritis & L40.5 \\
    & Systemic Lupus Erythematosus & M32 \\
    & Polymyalgia Rheumatica & M35.3, M31.5 \\
    & Mixed Connective Tissue Disease & M35.1 \\
    & Polymyositis/Dermatomyositis & M33 \\
    & Systemic Sclerosis & M34 \\
    & Fibromyalgia & M79.7 \\
\bottomrule
\end{tabular}
\caption{Phenotypes and associated \ac{ICD-10-CM} codes for the \ac{HPB} and rheumatic differential diagnosis tasks. All subcategories of the \ac{ICD-10-CM} codes here are also used. We considered a patient to have received the diagnosis if it appears at least twice in the patient's record on different dates. If an off-target diagnosis appears exactly once in a patient's record, that patient is excluded from analyses related to that off-target diagnosis.}
\label{table:diff_dx_definitions}
\end{table}

\begin{table}[htbp]
\centering
\tiny
\begin{tabular}{ m{3cm} m{12cm} }
\toprule
\textbf{Disease} & \textbf{Medical Codes for Disease-Specific Outcomes Tasks} \\
\midrule \ac{ASCVD}  & G45, I20.0, I20.8, I20.9, I21, I22, I24.9, I25.11, I25.7, I25.810, I25.812, I63, I70.2, I70.3, I70.4, I70.5, I70.6, I70.7, I70.8, I70.9, I73.9, I75.029, I77.6, Q28.8, Q87.89, T82.21, T82.310, T82.311, T82.312, T82.320, T82.321, T82.322, T82.330, T82.331, T82.332, T82.390, T82.391, T82.392, Z95.5\\
\midrule \ac{CKD} (Stage 2)  & N18.2 \\
\midrule \ac{CKD} (Stage 3)  & N18.3 \\
\midrule \ac{CKD} (Stage 4\textsuperscript{+})  & I12.0, I13.11, I13.2, N18.4, N18.5, N18.6 \\
\midrule Diabetic Neuropathy & E08.4, E10.4, E11.4, E13.4 \\
\midrule Diabetic Retinopathy & E08.31, E08.32, E08.33, E08.34, E08.35, E10.31, E10.32, E10.33, E10.34, E10.35, E11.31, E11.32, E11.33, E11.34, E11.35, E13.31, E13.32, E13.33, E13.34, E13.35 \\
\midrule Heart Attack & I21, I22 \\
\midrule Chronic Heart Failure  & I50 \\
\midrule Stroke & I60, I61, I63 \\
\midrule \ac{HgbA1c} & LOINC Codes: 4548-4, 4549-2, 17855-8, 17856-6 \\
\bottomrule
\end{tabular}
\caption{Associated codes for Disease-Specific Outcomes tasks. Subcategories of listed medical codes were included as part of the task definition. All medical codes are \ac{ICD-10-CM} unless otherwise specified.}
\end{table}

\begin{table}[htbp]
\centering
\tiny
\begin{tabular}{ m{3cm} m{1cm} m{11cm} }
\toprule
\textbf{Disease} & \textbf{Phenotype} & \textbf{ICD-10-CM Codes} \\
\midrule
Heart Attack\cite{ccw-chronic-conditions} & Emergent & I21.01, I21.02, I21.09, I21.11, I21.19, I21.21, I21.29, I21.3, I21.4, I21.9, I21.A1, I21.A9, I21.B, I22.0, I22.1, I22.2, I22.8, I22.9, I23.0, I23.1, I23.2, I23.3, I23.4, I23.5, I23.6, I23.7, I23.8 \\
\midrule
Alcohol Use Disorder*\cite{jaman2o24alc} & Chronic & F10.10, F10.120, F10.121, F10.129, F10.14, F10.150, F10.151, F10.159, F10.180, F10.181, F10.182, F10.188, F10.19, F10.20, F10.220, F10.221, F10.230, F10.231, F10.232, F10.239, F10.24, F10.250, F10.251, F10.259, F10.26, F10.27, F10.280, F10.281, F10.282, F10.288, F10.29 \\
\midrule
Alcohol Withdrawal \newline Syndrome*\cite{jaman2o24alc} & Emergent & F10.130, F10.131, F10.132, F10.139, F10.229, F10.230, F10.231, F10.232, F10.239, F10.930, F10.931, F10.932, F10.939 \\
\midrule
Asthma*\cite{ccw-chronic-conditions} & Chronic & J45.20, J45.21, J45.22, J45.30, J45.31, J45.32, J45.40, J45.41, J45.42, J45.50, J45.51, J45.52, J45.901, J45.902, J45.909, J45.990, J45.991, J45.998 \\
\midrule
Asthma Exacerbation*\textsuperscript{\dag}
\cite{ccw-chronic-conditions} & Emergent & J45.21, J45.22, J45.31, J45.32, J45.41, J45.42, J45.51, J45.52, J45.901, J45.990 \\
\midrule
\ac{COPD}*\cite{ccw-chronic-conditions} & Chronic & J40, J41.0, J41.1, J41.8, J42, J43.0, J43.1, J43.2, J43.9, J44.0, J44.1, J44.81, J44.89, J44.9, J47.0, J47.1, J47.9, J98.2, J98.3 \\
\midrule
\ac{COPD} Exacerbation*\textsuperscript{\dag}
\cite{ccw-chronic-conditions} & Emergent & J44.0, J44.1, J47.0, J47.1 \\
\midrule
\ac{CHF}*\cite{Graham2024phenotypes} & Chronic & I09.81, I11.0, I13.0, I13.2, I50.1, I50.20, I50.21, I50.22, I50.23, I50.30, I50.31, I50.32, I50.33, I50.40, I50.41, I50.42, I50.43, I50.810, I50.811, I50.812, I50.813, I50.814, I50.82, I50.83, I50.84, I50.89, I50.9 \\
\midrule
\ac{CHF} Exacerbation*\textsuperscript{\dag}
\cite{Graham2024phenotypes} & Emergent & I50.21, I50.23, I50.31, I50.33, I50.41, I50.43, I50.811, I50.813 \\
\midrule
Dementia\cite{ccw-chronic-conditions} & Chronic & F01.50, F01.51, F01.511, F01.518, F01.52, F01.53, F01.54, F01.A0, F01.A11, F01.A18, F01.A2, F01.A3, F01.A4, F01.B0, F01.B11, F01.B18, F01.B2, F01.B3, F01.B4, F01.C0, F01.C11, F01.C18, F01.C2, F01.C3, F01.C4, F02.80, F02.81, F02.811, F02.818, F02.82, F02.83, F02.84, F02.A0, F02.A11, F02.A18, F02.A2, F02.A3, F02.A4, F02.B0, F02.B11, F02.B18, F02.B2, F02.B3, F02.B4, F02.C0, F02.C11, F02.C18, F02.C2, F02.C3, F02.C4, F03.90, F03.91, F03.911, F03.918, F03.92, F03.93, F03.94, F03.A0, F03.A11, F03.A18, F03.A2, F03.A3, F03.A4, F03.B0, F03.B11, F03.B18, F03.B2, F03.B3, F03.B4, F03.C0, F03.C11, F03.C18, F03.C2, F03.C3, F03.C4, F05, G13.8, G31.01, G31.09, G31.1, G31.2, G31.83, G94, R41.81 \\
\midrule
Sickle Cell Crisis*\cite{admin2020peds} & Emergent & D57.0, D57.00, D57.01, D57.02, D57.03, D57.09, D57.21, D57.211, D57.212, D57.213, D57.218, D57.219, D57.41, D57.411, D57.412, D57.413, D57.414, D57.418, D57.419, D57.43, D57.431, D57.432, D57.433, D57.434, D57.438, D57.439, D57.45, D57.451, D57.452, D57.453, D57.454, D57.458, D57.459, D57.81, D57.811, D57.812, D57.813, D57.814, D57.818, D57.819 \\
\midrule
Sickle Cell Disease*\textsuperscript{\dag}
\cite{admin2020peds} & Chronic & D57, D57.0, D57.00, D57.01, D57.02, D57.03, D57.09, D57.1, D57.2, D57.20, D57.21, D57.211, D57.212, D57.213, D57.218, D57.219, D57.4, D57.40, D57.41, D57.411, D57.412, D57.413, D57.414, D57.418, D57.419, D57.42, D57.43, D57.431, D57.432, D57.433, D57.434, D57.438, D57.439, D57.44, D57.45, D57.451, D57.452, D57.453, D57.454, D57.458, D57.459, D57.8, D57.80, D57.81, D57.811, D57.812, D57.813, D57.814, D57.818, D57.819 \\
\bottomrule
\end{tabular}
\caption{Phenotypes and associated \ac{ICD-10-CM} codes for acute-on-chronic and incident disease risk prediction tasks. Asterisked phenotypes are used in acute-on-chronic tasks. Phenotypes with daggers are used for the acute definition of an acute-on-chronic task that we curated to be a subset of the cited chronic definition.}
\label{table:pop_aoc_definitions}
\end{table}

\clearpage
\section{Comparison of task-specific supervised models}
\begin{figure}[!htbp]
    \centering
    \includegraphics[width=0.85\linewidth]{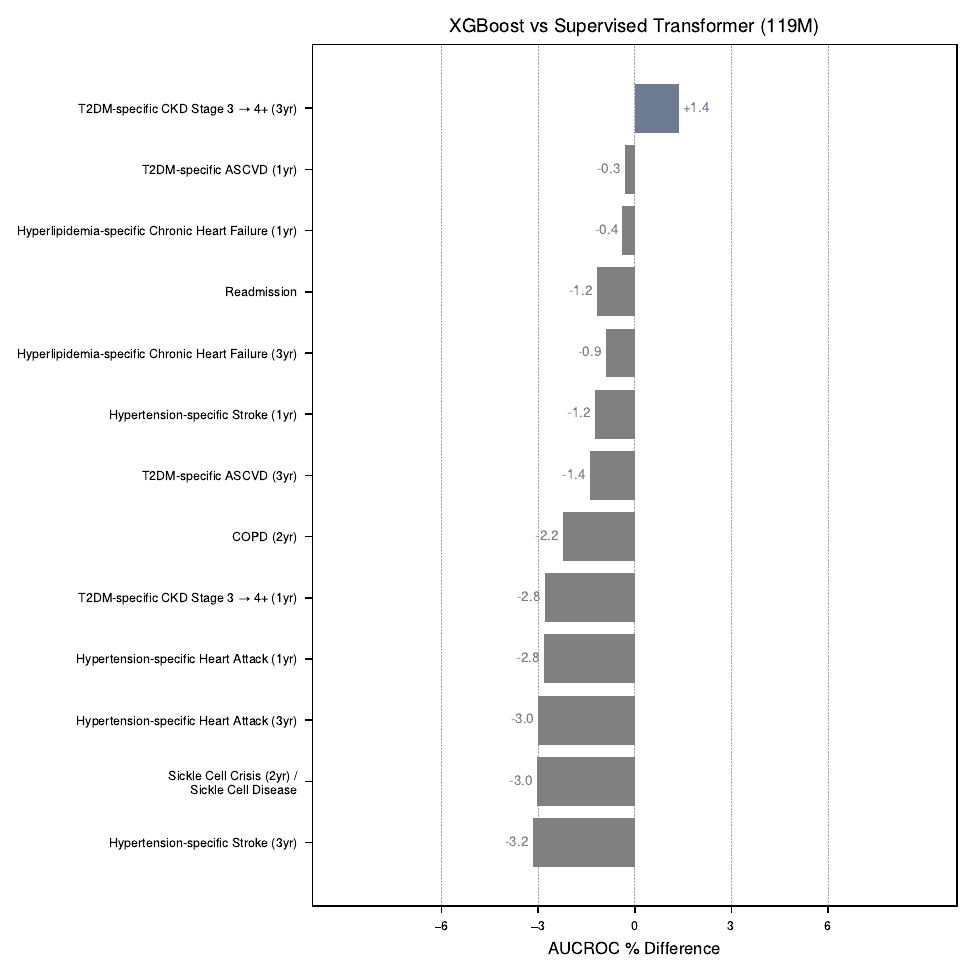}
    \caption[Comparative performance of task-specific supervised models.]{\textbf{XGBoost vs.\ Supervised Transformer (119M).} AUCROC comparisons between the supervised transformer and XGBoost models on a representative subset of classification tasks. Differences are shown in terms of percentage points with XGBoost performance as the baseline.}
    \label{fig:perf_diff_supervised_models}
\end{figure}

\end{document}